\documentclass{article} 
\usepackage{times}

\usepackage{iclr2024_conference}
\PassOptionsToPackage{sort,square,numbers}{natbib}
\usepackage{hyperref}       
\usepackage{url}     
\usepackage{booktabs}       
\usepackage{amsfonts}       
\usepackage{nicefrac}       
\usepackage{microtype}      
\usepackage{xcolor}         
\usepackage{enumitem} 
\usepackage{color, colortbl}
\usepackage{xcolor}
\usepackage{pifont}

\usepackage{graphicx}
\usepackage{epstopdf}
\usepackage{subcaption}
\usepackage{pgfplots}
\usepackage{pgfplotstable}
\usepackage{hyperref}
\usepackage{comment}
\usepackage{amsmath}
\usepackage{bm}
\usepackage{subcaption}
\usepackage{xcolor}

\usepackage[capitalise]{cleveref}
\usepackage{amssymb}
\usepackage{siunitx}
\usepackage{listings}
\usepackage[utf8]{inputenc}

\usepackage{booktabs}  
\usepackage{tabu}
\usepackage{array}
\usepackage{multirow}
\usepackage{diagbox}
\usepackage{amsthm}
\usepackage{wrapfig}
\usepackage{placeins}

\usepackage{algorithm,lipsum,changepage}
\usepackage{shortcuts_ln}

\usepackage{balance}

\usepackage{enumitem}
\usepackage{microtype}
\usepackage{calc}
\usepackage{ifthen}
\usepackage{hyphenat}
\usepackage{wrapfig}
\usepackage[font=small,labelfont=bf]{caption}
\usepackage{mathabx}
\usepackage{silence}

\usepackage{algpseudocode}
\usepackage{hyperref}

\newtheorem{definition}{Definition}[section]
\newcommand{\af}[0]{\Tilde{f}}

\begin{document}




\title{A General Framework for User-Guided Bayesian Optimization} 
\author{
  Carl Hvarfner \\
  Lund University\\
  \texttt{carl.hvarfner@cs.lth.se} 
  \And
  Frank Hutter \\
  University of Freiburg \\
  Prior Labs \\
  \texttt{fh@cs.uni-freiburg.de} \\
  \AND
  Luigi Nardi \\
  Lund University \\
  Stanford University \\
  DBTune \\
  \texttt{luigi.nardi@cs.lth.se} \\
}
\maketitle

\vskip 0.3in

\newcommand{\Ev}[0]{\mathbb{E}}
\newcommand{\ent}[0]{\text{H}}
\newcommand{\data}[0]{\mathcal{D}}
\newcommand{\hps}[0]{\bm{\theta}}
\newcommand{\yx}[0]{y_{\bm{x}}}
\newcommand{\fx}[0]{f_{\bm{x}}}
\newcommand{\mname}[0]{\texttt{ColaBO}}
\newcommand{\jes}[0]{\texttt{JES}}
\newcommand{\KG}[0]{\texttt{KG}}
\newcommand{\ei}[0]{\texttt{EI}}
\newcommand{\pimp}[0]{\texttt{PI}}
\newcommand{\logei}[0]{\texttt{LogEI}}
\newcommand{\pes}[0]{\texttt{PES}}
\newcommand{\es}[0]{\texttt{ES}}
\newcommand{\mes}[0]{\texttt{MES}}
\newcommand{\xopt}[0]{\bm{x}_*}
\newcommand{\fopt}[0]{f_*}
\newcommand{\opt}[0]{\bm{\bigast}}
\newcommand{\pibo}[0]{\texttt{$\pi$BO}}
\newcommand{\pfunc}[0]{P}

\newcommand{\frank}[1]{\todo[color=yellow]{Frank: #1}}
\newcommand{\erik}[1]{\todo[color=teal]{Erik: #1}}
\newcommand{\carl}[1]{{\todo[color=orange]{Carl: #1}}}
\newcommand{\luigi}[1]{\todo[color=olive]{Luigi: #1}}



\begin{abstract}
The optimization of expensive-to-evaluate black-box functions is prevalent in various scientific disciplines. Bayesian optimization is an automatic, general and sample-efficient method to solve these problems with minimal knowledge of the underlying function dynamics. However, the ability of Bayesian optimization to incorporate prior knowledge or beliefs about the function at hand in order to accelerate the optimization is limited, which reduces its appeal for knowledgeable practitioners with tight  budgets. To allow domain experts to customize the optimization routine, we propose \mname{}, the first Bayesian-principled framework for incorporating prior beliefs beyond the typical kernel structure, such as the likely location of the optimizer or the optimal value. The generality of \mname{} makes it applicable across different Monte Carlo acquisition functions and types of user beliefs. We empirically demonstrate \mname{}'s ability to substantially accelerate optimization when the prior information is accurate, and to retain approximately default performance when it is misleading. 

\end{abstract}

\section{Introduction} \label{intro}
\textit{Bayesian Optimization} (BO)~\citep{Mockus1978, jonesei, snoek-nips12a} is a well-established methodology for the optimization of expensive-to-evaluate black-box functions. Known for its sample efficiency, BO has been successfully applied to a variety of domains where laborious system tuning is prominent, such as hyperparameter optimization~\citep{snoek-nips12a,NIPS2011_86e8f7ab,lindauer_smac3}, neural architecture search~\citep{ru2021interpretable,white-aaai21a}, robotics~\citep{calandra-lion14a, mayr2022learning}, hardware design~\citep{nardi2019practical,ejjeh2022hpvm2fpga}, and chemistry~\citep{griffiths2020constrained}.

Typically employing a Gaussian Process~\citep{rasmussen-book06a} (GP) surrogate model, BO allows the user to specify a prior over functions $p(f)$ via the Gaussian Process kernel, as well as an optional prior over its hyperparameters. Within the framework of the prior, the user can specify expected smoothness, output range and possible noise level of the function at hand, with the hopes of accelerating the optimization if accurate. However, the prior beliefs that can be specified within the framework of the kernel hyperparameters do not span the full range of beliefs that practitioners may possess. For example, practitioners may know which \textit{parts of the input space} tend to work best~\citep{oh2018bock, Perrone2019LearningSS, smith2018disciplined, atmseer}, know a range or upper bound on the optimal output~\citep{pmlr-v139-jeong21a, pmlr-v119-nguyen20d} such as a maximal achievable accuracy of 100\%, or other properties of the objective, such as preference relations between configurations~\citep{huang2022bayesian}. The limited ability of practitioners to interact and collaborate with the BO machinery~\citep{NEURIPS2022_6751611b} thus runs the risk of failing to use valuable domain expertise, or alienating knowledgeable practitioners altogether. 
\begin{figure} 
\begin{minipage}{0.32\linewidth}
  \includegraphics[width=\linewidth]{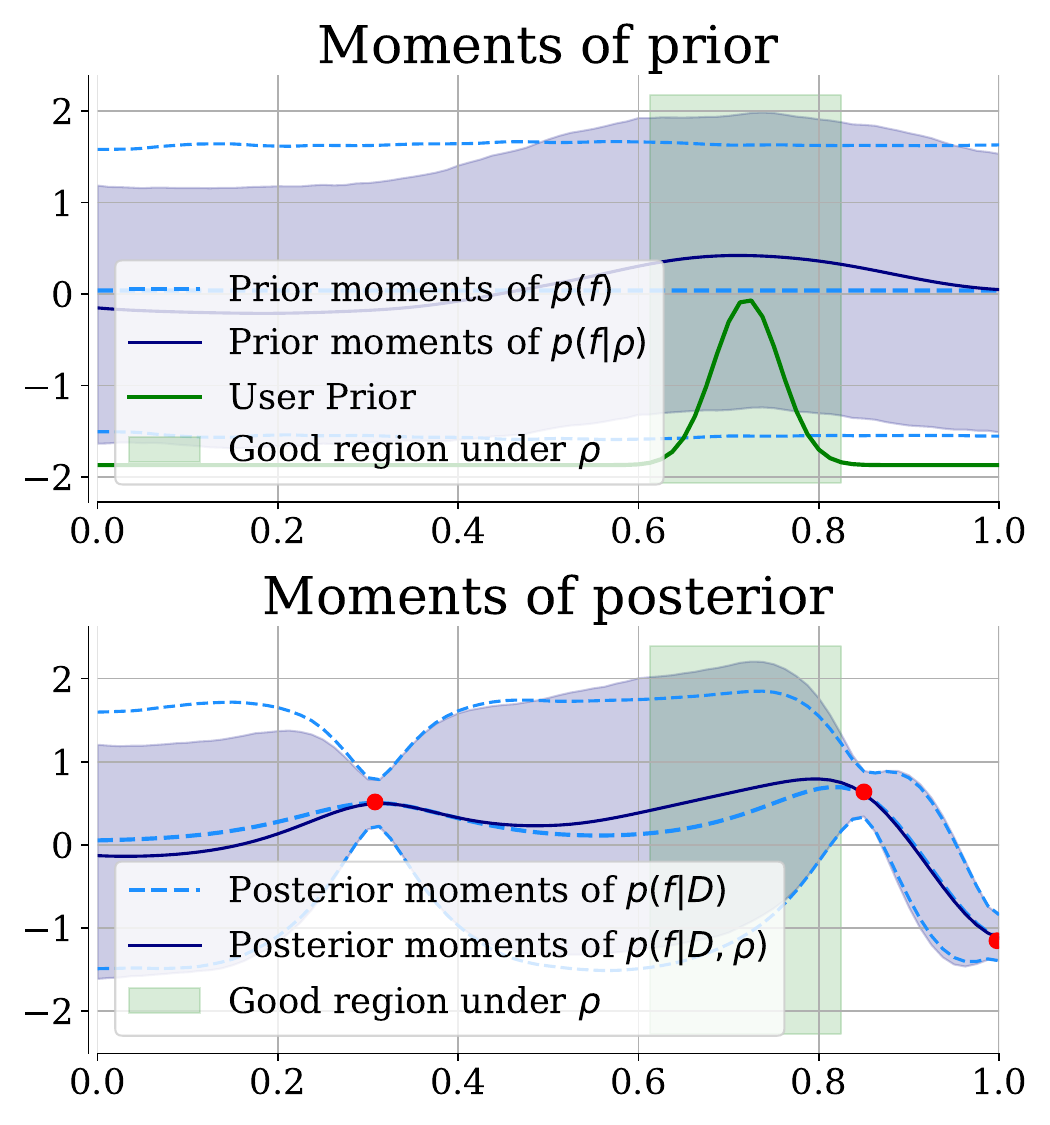}
\end{minipage}
\begin{minipage}{0.32\linewidth}
  \includegraphics[width=\linewidth]{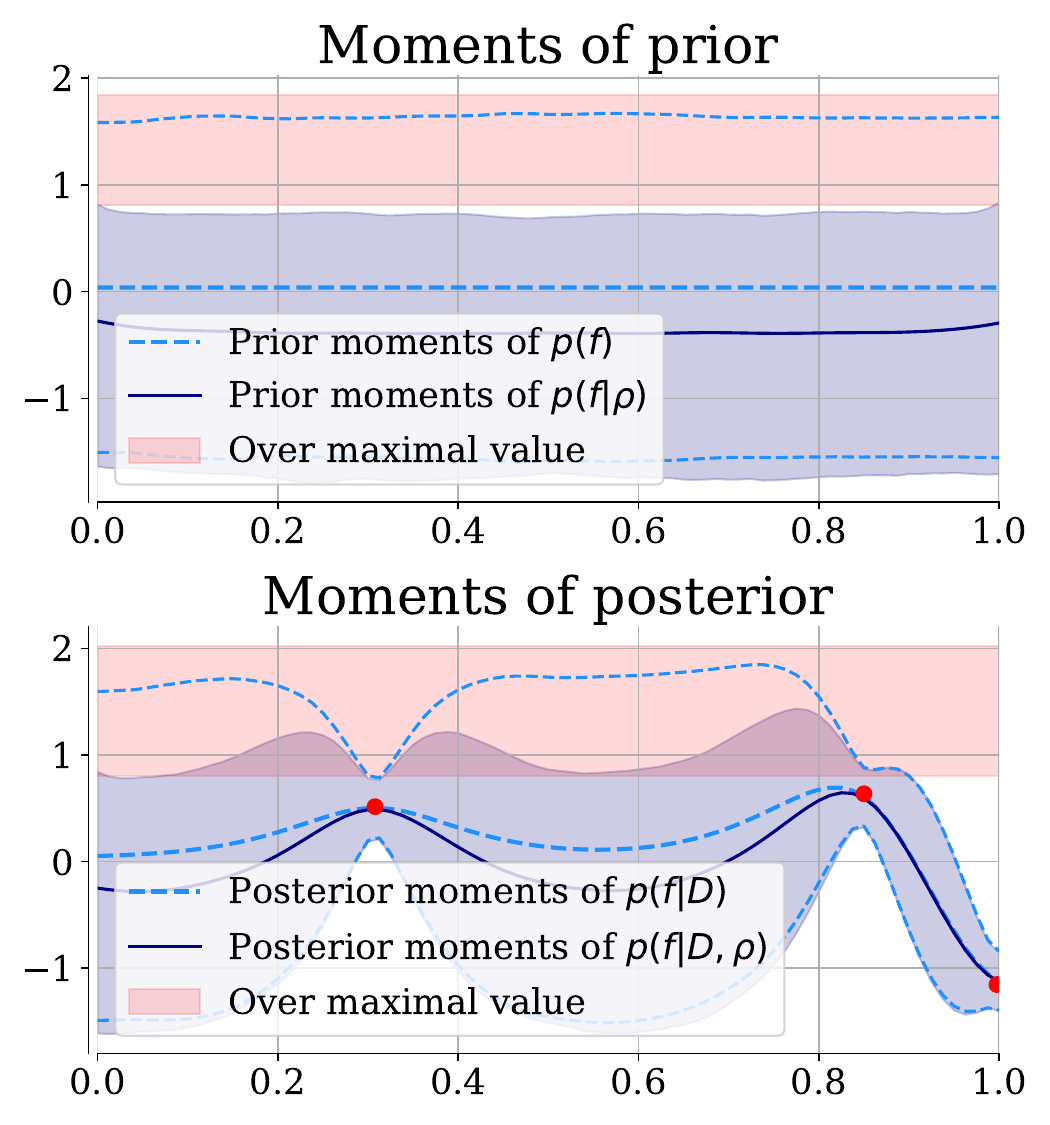}  
\end{minipage}
\begin{minipage}{0.32\linewidth}
  \includegraphics[width=\linewidth]{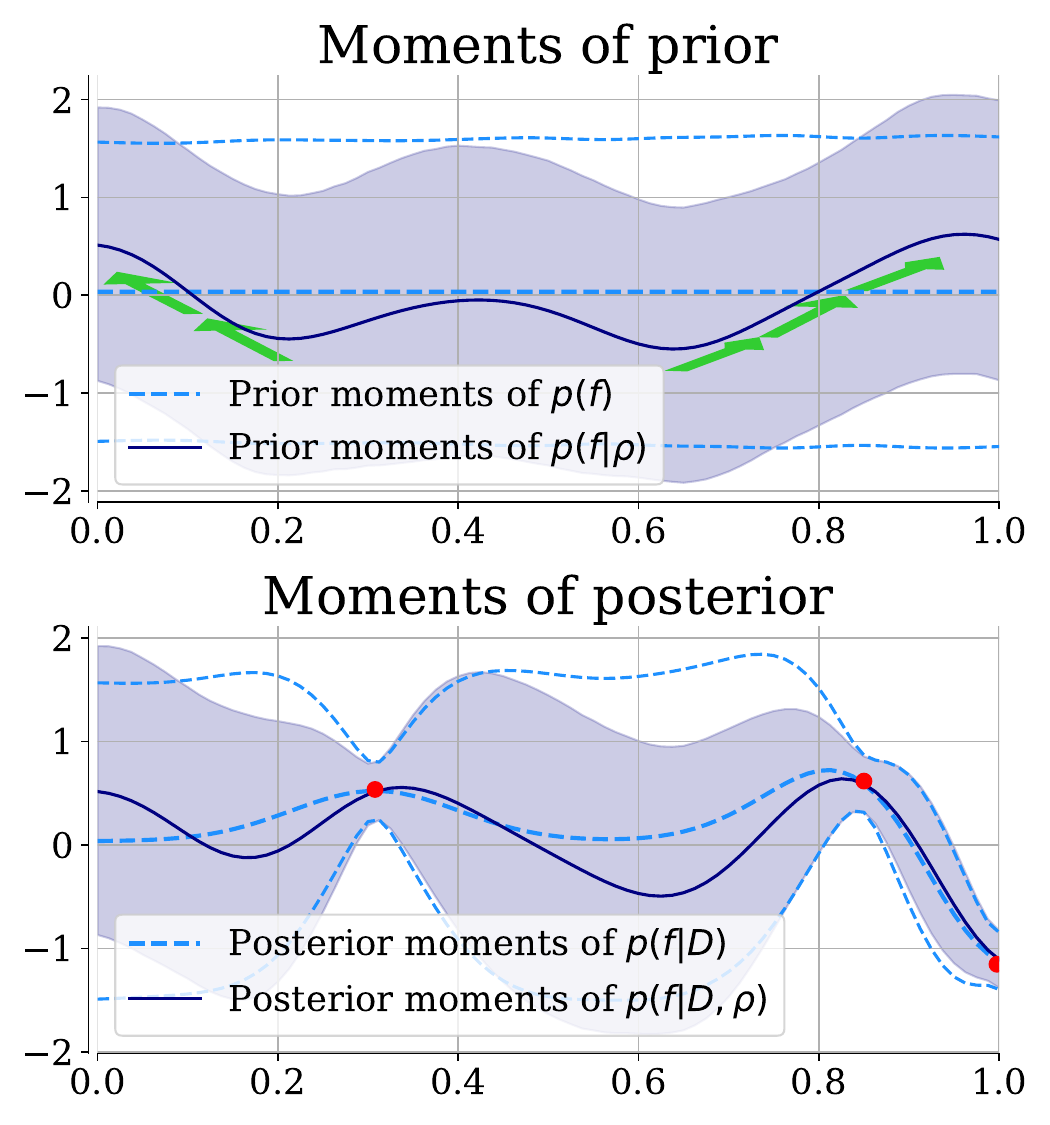}
\end{minipage}
\caption{Three different \mname{} priors: (left) Prior over the optimum $\xopt$, and the induced changed in the GP for an optimum located in the green region. (middle) Prior over optimal value, $f^* < 0.8$. (right) Prior over preference relations $f(\bm{x})_1 \geq f(\bm{x}_2)$ for five preferences (green arrows, e.g. $f(0.0) \geq f(0.1) \geq f(0.2)$.}
\vspace{-3mm}
\label{fig:intro}
\end{figure}
While knowledge injection beyond what is natively supported by the GP kernel is crucial to further increase the efficiency of Bayesian optimization, so far no current approach allows for the integration of arbitrary types of user knowledge. To address this gap, we propose an intuitive framework that effectively allows the user to reshape the Gaussian process at will to mimic their held beliefs. 

Figure~\ref{fig:intro} illustrates how, for the three aforementioned priors, the GP is reshaped to \textit{faithfully represent} the belief held by the user - whether it be a prior over the optimum (left), optimal value (middle), or preference relations between points (right). Our novel framework for \emph{Collaborative Bayesian Optimization} (\mname{}) diverges from the typical assumption of Gaussian posteriors, and is applicable to any Monte Carlo acquisition function~\citep{wilson2017reparam, NEURIPS2018_498f2c21, balandat2020botorch}. Formally, we make the following contributions:
\begin{enumerate}
    \item We introduce~\mname{}, a framework for incorporating  user knowledge over the optimizer, optimal value and preference relations into Bayesian optimization in the form of an additional prior on the surrogate, orthogonal to the conventional prior on the kernel hyperparameters, 
    \item We demonstrate that the proposed framework is generally applicable to Monte Carlo acquisition functions, inheriting MC acquisiiton function utility,
    \item We empirically show that \mname{} accelerates optimization when injected with for priors over optimal location and optimal value.
\end{enumerate}

\section{Background}\label{sec:background}
We outline Bayesian optimization, Gaussian Processes and Monte Carlo (MC) acquisition functions, as well as the concept of a prior over the optimum.
\subsection{Bayesian optimization}\label{sec:bo}
We consider the problem of optimizing a black-box function $f$ across a set of feasible inputs $\mathcal{X}\subset\mathbb{R}^d$:
\begin{equation}
    \bm{x}^* \in \argmax_{\bm{x}\in \mathcal{X}} f(\bm{x}).
    \label{eq:boo}
    \end{equation}
We assume that $f(\bm{x})$ is expensive to evaluate and can potentially only be observed through a noise-corrupted estimate, $y_{\bm{x}}$, where $y_{\bm{x}} = f(\bm{x}) + \epsilon, \epsilon \sim \mathcal{N}(0, \sigma_\epsilon^2)$ for some noise level $\sigma_\epsilon^2$. In this setting, we wish to maximize $f$ in an efficient manner.  Bayesian optimization (BO) aims to globally maximize $f$ by an initial design and thereafter sequentially choosing new points $\bm{x}_n$ for some iteration $n$, creating the data $\data{}_n = \data{}_{n-1} \cup \{{(\bm{x}_n, y_n)}\}$~\citep{brochu-arXiv10a,shahriari-ieee16a,garnett-book22a}. After each new observation, BO constructs a probabilistic surrogate model $p(f|\data_n)$~\citep{snoek-nips12a, hutter-lion11a, bergstra-nips11a, pmlr-v202-muller23a} and uses that surrogate to build an acquisition function $\alpha(\bm{x}; \data{}_n)$ which selects the next query.

\subsection{Gaussian processes}\label{sec:gp} 
When constructing the surrogate, the most common choice is a \textit{Gaussian process} (GP)~\citep{rasmussen-book06a}. The GP utilizes a covariance function $k$, which encodes a prior belief for the smoothness of $f$, and determines how previous observations influence prediction. Given observations $\data_n$ at iteration $n$, the Gaussian posterior $p(f|\data{}_n)$ over the objective is characterized by the posterior mean $\mu_n(\bm{x}, \bm{x}')$ and (co-)variance $\Sigma_n(\bm{x}, \bm{x}')$ of the GP:
\begin{align*}
    \mu_n(\bm{x}) &= \mathbf{k}_n(\bm{x})^\top(\mathbf{K}_n + \sigma_\epsilon^2\mathbf{I})^{-1}
    \mathbf{y}\\
    \Sigma_n(\bm{x}, \bm{x}') &=  k(\bm{x}, \bm{x}') - \mathbf{k}_n(\bm{x})^\top(\mathbf{K} + \sigma_\epsilon^2\mathbf{I})^{-1}\mathbf{k}_n(\bm{x}'),
\end{align*}
where $(\mathbf{K}_n)_{ij} = k(\bm{x}_i, \bm{x}_j)$, $\mathbf{k}_n(\bm{x}) = [k (\bm{x}, \bm{x}_1), \ldots, k(\bm{x}, \bm{x}_n)]^\top$ and $\sigma^2_\epsilon$ is the noise variance. 

For applications in BO and beyond, samples from the posterior are required either directly for optimization~\citep{eriksson2019scalable} through Thompson sampling~\citep{thompson1933likelihood}, or to estimate auxiliary quantities of interest~\citep{hernandez2015predictive, pmlr-v139-neiswanger21a, hvarfner2023selfcorrecting}. For a finite set of $k$ query locations $(\bm{X} = \bm{x}_1, \ldots, \bm{x}_k)$, the classical method
of generating samples is via a location-scale transform
of Gaussian random variables, $f(\bm{X}) = \mu_n(\bm{X}) + \bm{L}\bm{\epsilon}$, where $\bm{L}$ is the Cholesky decomposition of $\bm{K}$ and $\bm{\epsilon} \sim \mathcal{N}(0, \bm{I})$. Unfortunately, the classic approach is intrinsically non-scalable, incurring a $\mathcal{O}(k^3)$ cost due to the aforementioned matrix decomposition.

\subsection{Decoupled Posterior Sampling}\label{sec:decoupled}

To remedy the issue of scalability in posterior sampling, $\mathcal{O}(k)$ weight-space approximations based on Random Fourier Features (RFF)~\citep{rahimi2007rff} obtain approximate (continuous) function draws $\af(\bm{x}) = \sum_{i=1}^m w_i\phi_i(\bm{x})$, where $\phi_i(\bm{x}) = \frac{2}{\ell} (\bm{\psi}_i^\top \bm{x}+ b_i)$.  The random variables $w_i \sim \mathcal{N}(0, 1)$, $b_i \sim \mathcal{U}(0, 2 \pi)$, and $\bm{\psi}_i$ are sampled proportional to the spectral density of $k$.

While achieving scalability, the seminal RFF approach by \citet{rahimi2007rff} suffers from the issue of variance starvation~\citep{mutny18efficient, pmlr-v84-wang18c, wilson2020efficiently}. As a remedy,~\citet{ wilson2020efficiently} decouple the draw of functions from the approximate posterior $p(\af|\data{})$ into a more accurate draw from the prior $p(\af)$, followed by a deterministic data-dependent update:
\begin{equation}\label{eq:pathwise}
    (\af|\data{})(\bm{x}) \,{\buildrel d \over =}\, \underbrace{\af(\bm{x})}_\text{draw from prior} + \;\;\;
    \underbrace{\mathbf{k}_n(\bm{x})^\top(\mathbf{K}_n + \sigma_\epsilon^2\mathbf{I})^{-1}
    (\mathbf{y} - \af(\bm{x}) - \bm{\epsilon})}_\text{deterministic update}
\end{equation}
Eq.~\ref{eq:pathwise} deviates from the distribution-first approach that is typically prevalent in GPs in favor of a variable-first approach utilizing Matheron's rule~\citep{etde_5214736}.

\subsection{Monte Carlo Acquisition Functions}\label{sec:myopic}
Acquisition functions act on the surrogate model to quantify the utility of a point in the search space. They encode a trade-off between exploration and exploitation, typically using a greedy heuristic to do so. A simple and computationally cheap heuristic is Expected Improvement (\ei{})~\citep{jonesei, bull-jmlr11a}. For a noiseless function and a current best observation $y_n^*$,  the \ei{} acquisition function is $\alpha_{EI}(\bm{x}) = \mathbb{E}_{\yx{}}\left[(y_n^* - \yx{})^+\right]$. For noisy problem settings, a noise-adapted variant of \ei{}~\citep{letham-ba18a} is frequently considered, where both the  incumbent $y_n^*$ and the upcoming query $\yx{}$ are substituted for the non-observable noiseless incumbent $f_n ^*$ and noiseless upcoming query $f_{\bm{x}}$. 
Other frequently used acquisition functions are the Upper Confidence Bound (\texttt{UCB})~\citep{Srinivas_2012},  Probability of Improvement (PI)~\citep{pi} and Knowledge Gradient (KG) ~\citep{frazier2009knowledge}. 
Information-theoretic acquisition functions consider the mutual information to select the next query  $\alpha_{\texttt{MI}}(\bm{x}) = I((\bm{x}, \yx{}); \bm{*}|\data_n)$, 
where $\bm{*}$ can entail either the optimum $\xopt$ as in (Predictive) Entropy Search (\es{}/\pes{})~\citep{entropysearch, pes}, the optimal value $\fopt$ as in Max-value Entropy Search (\mes{})~\citep{wang2017maxvalue,moss2021gibbon} or the tuple $(\xopt, \fopt)$ for Joint Entropy Search (\jes{})~\citep{hvarfner2022joint, tu2022joint}. 

All the aforementioned acquisition functions compute expectations $\mathbb{E}_{f_{\bm{x}}}$ (or alternatively $\mathbb{E}_{\yx{}}$) over some utility $u(f_{\bm{x}})$ of the output~\citep{wilson2017reparam, NEURIPS2018_498f2c21}, which typically have simple, or even closed-form, solutions for Gaussian posteriors. However, approximating the expectation through Monte Carlo integration has proven useful in the context of batch optimization~\citep{NEURIPS2018_498f2c21}, efficient acquisition function approximation~\citep{balandat2020botorch}, and non-Gaussian posteriors~\citep{astudillo2021bayesian}. By sampling over possible outputs $f_{\bm{x}}$ and utilizing the reparametrization trick~\citep{vae, pmlr-v32-rezende14}, utilities $u$ can be easily computed across a larger set of applications and be optimized to greater accuracy.

\subsection{Prior over the Optimum}\label{sec:prior} 
 A prior over the optimum~\citep{souza2021bayesian,hvarfner2022pibo, mallik2023priorband} is a user-specified belief $\pi: \mathcal{X} \to \mathbb{R}$ of the subjective likelihood that a given $\bm{x}$ is optimal. Formally,
\begin{equation}\label{eq:prior}
    \pi(\bm{x}) = \mathbb{P}\left(\bm{x} = \argmax_{\bm{x'}} f(\bm{x'})\right).
\end{equation}
This prior is generally considered to be independent of observed data, but rather a result of previous experimentation or anecdotal evidence. Regions that the user expects to contain the optimum will typically have a high value, but this does not exclude the chance of the user belief $\pi (\bm{x})$ to be inaccurate, or even misleading. Lastly, we require $\pi$ to be strictly positive in all of $\mathcal{X}$, which suggests that any point included in the
search space may be optimal.

\section{Methodology}
We now introduce \mname, the first Bayesian-principled BO framework that flexibly allows users to \textit{collaborate} with the optimizer by injecting prior knowledge about the objective that substantially exceeds the type of prior knowledge natively supported by GPs. In Sec.~\ref{sec:hierarchy}, we introduce and derive a novel prior over function properties, which yields a surrogate model conditioned on the user belief. Thereafter, in Sec.~\ref{sec:mcacq}, we demonstrate how the hierarchical prior integrates with MC acquisition functions. Lastly, in Sec.~\ref{sec:practical_considerations}, we state practical considerations to assure the performance of \mname.

\subsection{Prior over Function Properties}\label{sec:hierarchy}

We consider the typical GP prior over functions $p(f) = \mathcal{GP}(\mu, \Sigma)$, where the characteristics of $f$, such as smoothness and output magnitude, are fully defined by the kernel $k$ (and its associated hyperparameters $\hps{}$, which are omitted for brevity). We seek to inject an additional, user-defined prior belief over $f$ into the GP, such as the prior over the optimum in Sec.~\ref{sec:prior}, $\pi(\bm{x}) = \mathbb{P}\left(\bm{x} = \argmax_{\bm{x'}} f(\bm{x'})\right)$.  By postulating that $\pi$ is accurate, we wish to form a belief-weighted prior - a prior over \textit{functions} where the distribution over the optimum is exactly $\pi(\bm{x})$.
We start by considering the user belief $\pi: \mathcal{X} \xrightarrow{} \mathbb{R}$ from Eq.~\eqref{eq:prior}, and extend the definition to involve the integration over $f$, similarly to the Thompson sampling definition of~\cite{kandasamy-aistats18a}. Formally,
\begin{equation} \label{eq:pidirac}
    \pi(\bm{x}) = \mathbb{P}\left(\bm{x} = 
    \argmax_{\bm{x'}} f(\bm{x'})\right) = \int_f \pi(\delta_*(\bm{x}|f)) p(f)df
\end{equation}
where $\delta_*(\bm{x}|f) = 1, \text{if } \bm{x} = \argmax_{\bm{x}' \in \mathcal{X}} f(\bm{x}')$, and zero otherwise. As such, $\delta_*(\bm{x}|f)$ maps a function $f_i \sim p(f)$ to its $\argmax$, and evaluates whether this $\argmax$ is equal to $\bm{x}$. 

However, a belief over the optimum, or any other property, of a function $f$ is implicitly a belief over the function $f$ itself. As such, a non-uniform $\pi(\bm{x})$ should reasonably induce a change in the prior $p(f)$ to reflect the non-uniform optimum. 
To this end, we introduce an augmented user belief over the optimum $\rho_{\bm{x}}^*\sim \mathcal{P}_{\bm{x}}^*$, where $\mathcal{P}_{\bm{x}}^*$ is the prior over possible user beliefs, and draws are random functions $\rho_{\bm{x}}^*: \mathcal{X} \rightarrow \mathbb{R}^+$
which themselves take a function $f$ as input, and output a positive real number quantifying the likelihood of a sample $f_i$ under $\pi(\bm{x})$. Formally, we define $\rho_{\bm{x}}^*$ as
\begin{equation}\label{eq:rhodef}
  \rho_{\bm{x}}^*(f)  = \mathbb{P}\left(\bm{x} = 
    \argmax_{\bm{x'}} f(\bm{x'})\right) = \frac{1}{Z_{\rho_{\bm{x}}^*}} \int_\mathcal{X} \delta_*(\bm{x}|f) \pi(\bm{x})d\bm{x}
\end{equation}
where the intractible normalizing constant $Z_{\rho_{\bm{x}}^*}$ arises from the fact that the integrated density $\pi(\bm{x})$ acts on a finite-dimensional \textit{property} of $f$, and not $f$ itself. Under $\rho_{\bm{x}}^*(f)$, functions whose $\argmax$ lies in a high-density region under $\pi$ will be assigned a higher probability. 
Notably, the definition in~\ref{eq:rhodef} can extend to other properties of $f$ as well: a user belief $p_{f_*}$ over the optimal value $f_*$ analogously yields
a belief over functions $\rho_{f_{\bm{x}}}^*(f)$: 
\begin{equation}\label[type]{eq:rhooptval}
  {\rho_{f_{\bm{x}}}^*}(f)  = \mathbb{P}\left({\bm{x}} = 
    \max_{\bm{x'}} f(\bm{x'})\right) = \frac{1}{Z_{\rho_{f_{\bm{x}}}^*}}\int_{f_{\bm{x}}} \delta_*(\bm{x}|f) p_{f^*}(f_{\bm{x}})df_{\bm{x}}.
\end{equation}
Notably, we integrate over $f_{\bm{x}}$ (and not $y_{\bm{x}}$) to signify that the optimal function value does not involve observation noise~\cite{pmlr-v119-takeno20a, pmlr-v162-takeno22a}. It is worthwhile to reflect on the meaning of $\rho(f)$, and how beliefs over function properties propagate to $p(f)$. 
Concretely, if the user belief ${\rho_{f_{\bm{x}}}^*}(f)$ asserts that the maximal value lies within $C_1  < \max f < C_2$, the resulting distribution over $f$ should only contain functions whose $\max$ falls within this range. 
Using rejection sampling, functions which disobey this criterion are filtered out, which yields the posterior $p(f|\rho)$.
Having defined and exemplified how user beliefs impact the prior over functions $p(f)$, the role of 
$\rho$ as a likelihood should be apparent: given a prior over functions $p(f)$ and a user belief over functions $\rho(f)$ which places a probability
on all possible draws $f_i ~p(f)$, we can form a belief-weighted prior $p(f|\rho) \propto p(f)\rho(f)$. Thus, we introduce the formal definition of a user belief over a function property:
\begin{definition}[User Belief over Functions]
    The user belief over functions $\rho(f) \propto \frac {p(f|\rho)}{p(f)}$. 
\end{definition}
As the subsequent derived methodology applies
independently of the specific property of $f$ that a prior is placed on, we will henceforth consider a belief over a general function property $\rho$. Having defined the role of $\rho$ and the posterior over functions it produces, a natural question arises: \textit{How is $p(f|\rho)$ updated once observations $\data{}$ are obtained? }

Since the data $\data{}$ is independent of the prior (the data generation process is intrinsically unaffected by the belief held by the user), application of Bayes' rule yields the following posterior $p(f|\data,\rho$),
\begin{equation}\label{eq:bayes}
    p(f|\data{}, \rho) = \frac{p(\data, \rho|f)p(f)}{p(\data, \rho)} = \frac{p(\data|f)p(\rho|f)p(f)}{p(\data)p(\rho)} = \frac{p(f|\rho)}{p(f)} p(f|\data) \;\propto \;\rho(f)p(f|\data),
\end{equation} 

\begin{figure}[tbp]
  \centering
\includegraphics[width=\linewidth]{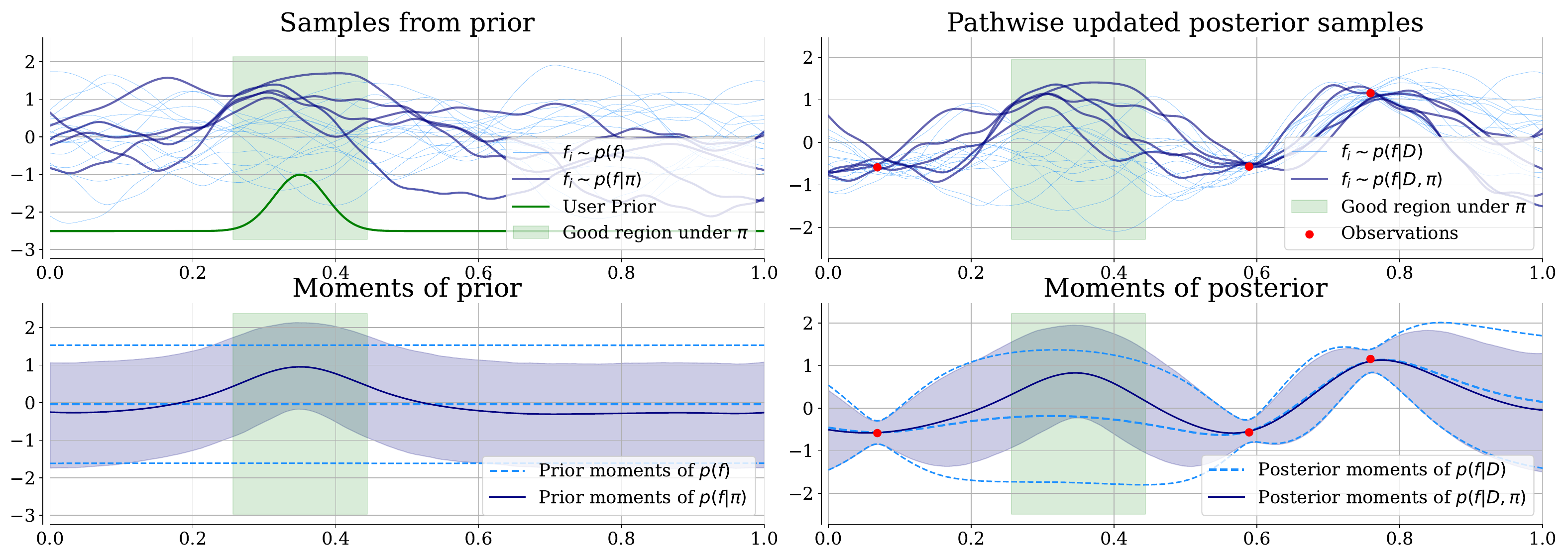} 
\caption{(Top left) Draws from the prior $p(f)$ (light blue) and the belief-weighted prior $p(f|\rho)$ whose members are likely to have their optimum within the green region. (Top right) Pathwise updated draws based on observed data. As the green region is distant from the observed data, samples are almost unaffected by the data in this region. (Bottom left) Exact mean and standard deviation ($\mu_{\bm{x}}, \sigma_{\bm{x}}$) of $p(f)$ and estimated mean and standard deviation of $p(f|\rho)$. (Bottom right) Exact $p(f|\data{})$ and estimated $p(f|\rho, \data)$. As $p(f|\rho)$ constitutes of functions whose optimum is located within the green region the resulting model has a higher mean and lower variance within this region. Moreover, $p(f|\rho)$ globally displays lower upside variance compared to the vanilla GP.}
\label{fig:belief_x}
\end{figure}
where the right side of the proportionality in Eq.~\ref{eq:bayes} suggests an intuitive generation process for samples $(f|\data{}, \rho)$ to approximate the density $p(f|\data{}, \rho)$. Utilizing the pathwise update from Eq.~\ref{eq:pathwise}, we note that given an approximate draw $\af{}$ from the prior, the subsequent data-dependent update is deterministic. Recalling Eq.~\ref{eq:pathwise} and assuming independence between $\rho$ and $\data{}$, $\rho$ only affects the draw from the prior, whereas $\data{}$ only affects the update. Consequently, we obtain
\begin{equation}\label{eq:pi_pathwise}
    (\af|\data{}, \rho)(\bm{x}) \,{\buildrel d \over =}\, \underbrace{(\af| \rho)(\bm{x})}_\text{draw from prior} + \;\;\;
    \underbrace{\mathbf{k}_n(\bm{x})^\top(\mathbf{K}_n + \sigma_\epsilon^2\mathbf{I})^{-1}
    (\mathbf{y} - (\af| \rho)(\bm{x}) - \bm{\epsilon})}_\text{deterministic update},
\end{equation}

where $(\af{}|\rho) \sim p(f)\rho(\af{})$ are once again obtained using rejection sampling on draws from $p(\af{})$. Figure~\ref{fig:belief_x} displays this in detail: given the typical GP prior over functions \textit{and} a user belief over the optimum, we obtain a distribution over functions $p(\af{}|\rho_{\bm{x}}^*)$ before having observed any data (top right). Samples from the approximate prior $p(\af{})$ (light blue) are  re-sampled proportionally to their probability of occurring under the prior $\rho_{\bm{x}}^*(\af{})$ in green, leaving samples $(\af{}|\rho_{\bm{x}}^*)$ in navy blue, which are highly probable under $\rho_{\bm{x}}^*$. Once data is obtained, these samples are updates according to Eq.~\ref{eq:pi_pathwise}, which preserves the shape of the samples far away from observed data and yields the desired posterior.

\subsection{Prior-weighted Monte Carlo Acquisition Functions}\label{sec:mcacq}
\begin{wrapfigure}[15]{r}{0.46\linewidth}
  \centering
  \vspace{-5mm}
\includegraphics[width=\linewidth]{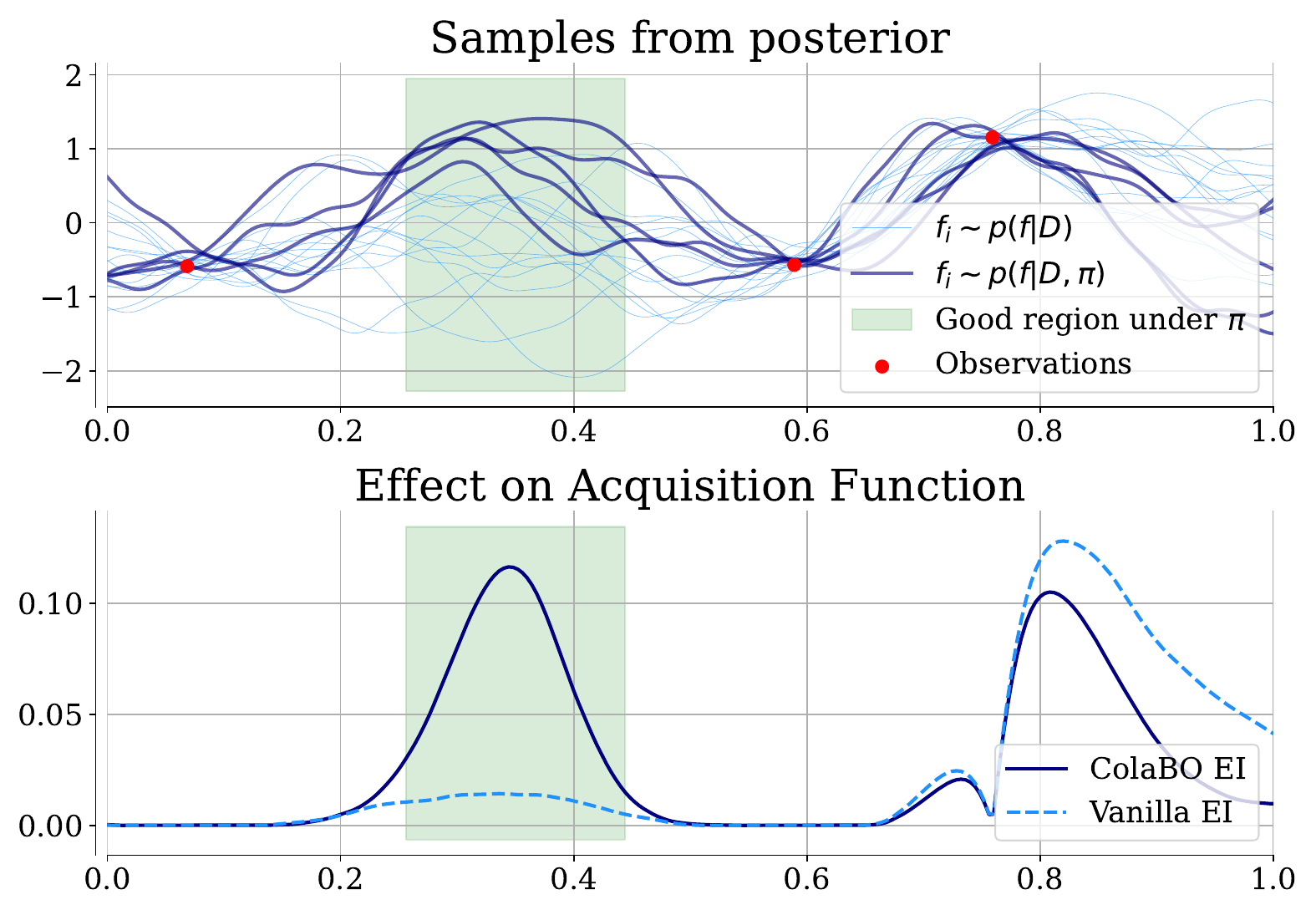} 
\caption{(Top) Draws from $p(f|\data{})$ (light blue) and $p(f|\rho, \data{})$ with a prior $\rho$ located in the green region. (Bottom) Vanilla MC-\ei{} and \mname{} MC-\ei{}, resulting from computing the acquisition function from sample draws from $p(f|\rho, \data{})$.}\label{fig:colaboei}
\end{wrapfigure}
Naturally, neither the belief-weighted prior $p(f|\rho)$ nor the belief-weighted posterior $p(f|\data, \rho)$ have a closed-form expression. Both are inherently non-Gaussian for non-uniform beliefs. As such, we resort to MC acquisition functions to compute utilities that are amenable to BO. In the subsequent section, we focus on the prevalent acquisition functions \ei{}, and \mes{}.

\paragraph{Expected Improvement} The computation of the MC-\ei{} within the \mname{} framework requires only minor adaptations of the original MC acquisition function. By definition, MC-\ei{} assigns utility $u$ as $u_\ei{}(f(\bm{x})) = \max(f^*_n - f(\bm{x}), 0)$, which yields
\begin{align}
    &\alpha_\ei{}(\bm{x}; \data{}) =  \mathbb{E}_{\fx|\data{}}[u_\ei{}(\fx)] \approx \\
    &\sum_\ell \max(f^*_n - \fx^{(\ell)}, 0),\; \fx^{(\ell)} \sim p(f(\bm{x})|\data{}).
\end{align}
Utilizing rejection sampling, we can compute the MC-\ei{} under the \mname{} posterior accordingly,
\begin{align}
&\alpha_\ei{}(\bm{x}; \data{}, \rho) = \mathbb{E}_{\fx|\data{},  \rho}[u_\ei{}(\fx)] \propto \\ 
&\int_f u_{\ei{}}(\fx)\rho(f) p(f|\data{})df \approx \sum_\ell \rho(f^{(\ell)}) \max(f^*_n - \fx^{(\ell)}, 0),\quad \fx^{(\ell)} \sim p(f(\bm{x})|\data{}), \label{eq:colaboei}
\end{align}
wherein samples in Eq.~\ref{eq:colaboei} are drawn from the prior, retained with probability $\rho(f^{(\ell)}) / \max \rho$, and pathwise updated. In Figure~\ref{fig:colaboei}, we demonstrate how \mname{}-\ei{} differs from MC-\ei{} for an identical posterior as in Figure~\ref{fig:belief_x}. By computing $\alpha_\ei{}$ from samples biased by $\rho$, \mname{} substantially directs the search towards good regions under $\rho$. Derivations for \pimp{} and \KG{} are analogous to that of \ei{}.

\paragraph{Max-Value Entropy Search} We derive a \mname{}-\mes{} acquisition function by first considering the definition of the entropy, $\ent[p(\yx|\data)] = \Ev_{\yx|\data}[-\log p(\yx|\data)]$. When considering the belief-weighted posterior, we further condition the posterior on $\rho$ and obtain
\begin{align}
    \label{eq:pesreform}
    &\small{\alpha_{\mes{}}(\bm{x}) =
    \Ev_{\fopt{}|\data, \rho} \left[\Ev_{\yx|\data, \rho, \fopt{}}[\log p(\yx|\data, \rho, \fopt{})]\right]}  - \Ev_{\yx|\data, \rho}[\log p(\yx|\data, \rho)] \\
    &\small{\propto\; \Ev_{\fopt{}|\data, \rho} \left[\Ev_{\fx|\data, \rho}[\Ev_{\yx|\fx}[\log p(\yx|\fx, \rho, \fopt)]]\right]} - \Ev_{\fx|\data, \rho}[\Ev_{\yx|\fx}[\log p(\yx|\fx, \rho)]] \\
    &\approx \frac{1}{Z_J}\sum_{j=1}^J \sum_{\ell=1}^L \sum_{k=1}^K \log p(y^{(k)}_{\bm{x}}|f^{(\ell)}_{\bm{x}}, {f^{(j)}_*})\rho(f^{(\ell)})\rho(f^{(j)}) - \sum_{\ell=1}^L \sum_{k=1}^K \log p(y^{(k)}_{\bm{x}}|f^{(\ell)}_{\bm{x}})\rho(f^{(\ell)}),\label{eq:mesfinal}
\end{align}
where $Z_J$ is a normalizing constant $\sum_J \rho(f^{(j)})$ brought on by sampling optimal values, $\yx |\fx$ can trivially be obtained by sampling Gaussian noise $\varepsilon\sim \mathcal{N}(0, \sigma_\varepsilon^2)$ to a noiseless observation $f_{\bm{x}} |\data$ in the innermost expectation, and $\fx$ and $\fopt$ are obtained through the pathwise sampling procedure outlined in Eq.~\ref{eq:pi_pathwise}. The samples are evaluated on $p((\yx|\fx)$,$(\yx|\fx, \fopt))$. As evident by Eq.~\ref{eq:mesfinal}, $\rho$ affects the posterior distribution of both the observations $\yx$ and the optimal values $\fopt$. \pes{} and \jes{} are derived analogously. However, these acquisition function require conditioning on additional, simulated data and consequently, additional pathwise updates, to compute.

\begin{algorithm}[tb] 
	\caption{\mname{} iteration\label{alg:colabo}} 
	\begin{algorithmic}[1]
	\State {\bfseries Input:} User prior $\rho$, number of function samples $L$, current data $\mathcal{D}$
	\State {\bfseries Output:} Next query location $\bm{x}'$.
	         \For{$\ell \in \{{1,\ldots, L}\}$}
            \State {$\rho^{(\ell)} = \rho(\af{}^{(\ell)}; 
             n), \af{}^{(\ell)} \sim p(\af)$} \algorithmiccomment{\small{\texttt{{Sample functions and evaluate on $\pi$}}}}
            \State {$(\af{}^{(\ell)}|\data{}) = \texttt{PathwiseUpdate}(\af{}^{(\ell)}, \data)$} \algorithmiccomment{\small{\texttt{{Per-sample update as in Eq.~\ref{eq:pi_pathwise}}}}}
        \EndFor
        \State {$p(\af|\data, \rho) \approx \sum_\ell \rho^{(\ell)} (\af{}^{(\ell)}|\data{})$}\algorithmiccomment{\small{\texttt{{Form MC estimate of posterior}}}}
      \State $\bm{x}' = \argmax_{\bm{x} \in \mathcal{X}} \mathbb{E}_{p(\af|\data, \rho)}[u(\af_{\bm{x}})]$ \;\;\;\algorithmiccomment{\small{\texttt{Maximize MC acquisition}}}
\end{algorithmic} 
\vspace*{-0.1cm}
\end{algorithm}

\subsection{Practical Considerations}
\label{sec:practical_considerations}\mname{} introduces additional flexibility to MC-based BO acquisition functions. The \mname{} framework deviates from vanilla (q-)MC acquisition functions~\citep{wilson2017reparam, balandat2020botorch} by utilizing approximate sample functions from the posterior, as opposed to pointwise draws from the posterior predictive and the reparametrization trick~\citep{pmlr-v32-rezende14}. \mname{} holds three shortcomings not prevalent in vanilla MC acquisition functions: (1) it cannot utilize Quasi-MC in the draws from the predictive posterior (only in the RFF weights), (2) it cannot fix the base samples~\citep{balandat2020botorch} drawn from the posterior for acquisition function consistency across the search space, and (3) the RFF approximation of $p(f)$ introduces bias. This approximation error is more pronounced for the Matérn 5/2-kernel than the squared exponential, leaving ColaBO best suited for the latter. In Sec.~\ref{sec:vanilla}, we display the impact of these shortcomings. While acquisition function optimization no longer enjoys the improved accuracy that stems from the reparametrization trick, the high degree of smoothness of function samples still allow for efficient gradient-based optimization.

\section{Results}
We evaluate the performance of \mname{} on various tasks, using priors over the optimum $\rho_{\xopt}$ obtained from known optima on synthetic tasks, as well as from prior work~\citep{mallik2023priorband} on realistic tasks. We consider two variants of~\mname{}: 
one using \logei{}~\citep{anonymous2023unexpected}, a numerically stable, smoothed $\texttt{logsumexp}$ transformation of~\ei{} with analogous derivation, and one variant using~\mes{}. We benchmark against the vanilla variants of each acquisition function, as well as \pibo{}~\citep{hvarfner2022pibo} and decoupled Thompson sampling~\cite{thompson1933likelihood, wilson2020efficiently}. 
All acquisition functions are implemented in BoTorch~\citep{balandat2020botorch} using a squared exponential kernel and MAP hyperparameter estimation. We present experiments with a Matérn-5/2~\citep{matern1960spatial} kernel in App.~\ref{sec:matern}. Unless stated otherwise, all methods are initialized with the mode of the prior followed by 2 Sobol samples. The experimental setup is presented in Appendix \ref{app:setup}, and our code is publicly available at \url{https://github.com/hvarfner/colabo}.

\subsection{Approximation Quality of the \mname{} Framework}~\label{sec:vanilla}
\begin{wrapfigure}{r}{0.48\linewidth}
  \centering
  \vspace{-7mm}
\includegraphics[width=\linewidth]{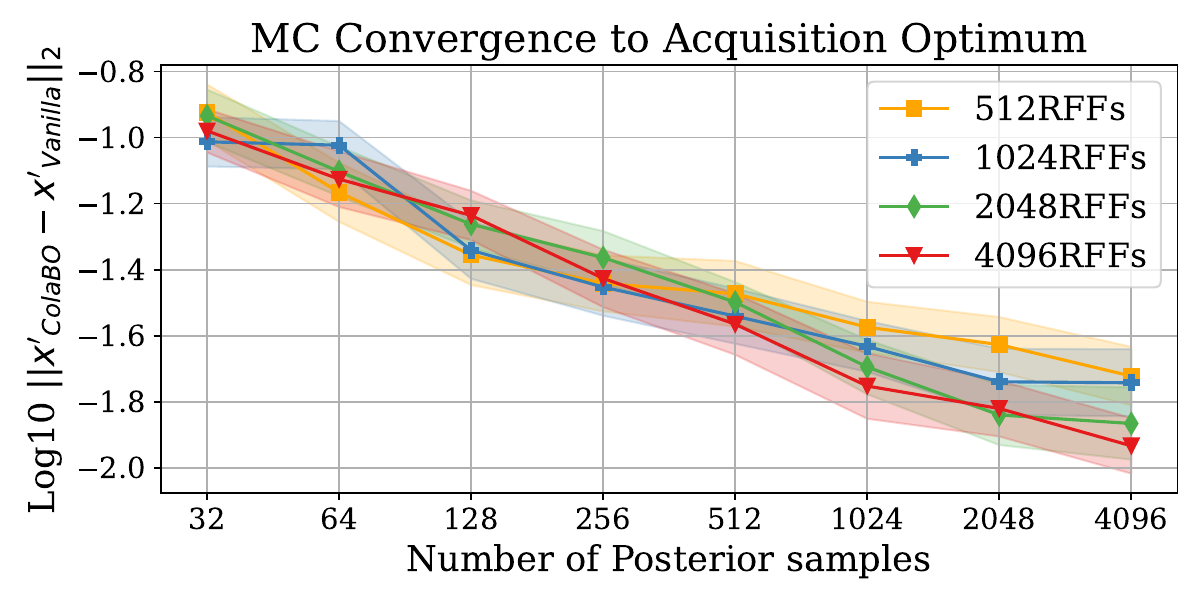} 
\caption{Mean and $1/4$ standard deviation of MC-induced errors of \mname{}-\logei{} relative vanilla \logei{} as measured by the distance to the $\argmax$ of the acquisition function on Hartmann (3D) on 10 randomly sampled points for 40 seeds.}\label{fig:errors}
  \vspace{-8mm}
\end{wrapfigure}
Firstly, we demonstrate the approximation quality of \mname{} \textit{without} user priors to assert its accuracy compared to a vanilla MC acquisition function. To facilitate comparison, we randomly sample 10 points on the Hartmann (3D) function, and optimize \logei{} with a large budget. We subsequently optimize \mname{}-\logei{} on the same set of points and compare the $\argmax$ to the solution found by the gold standard. Figure~\ref{fig:errors} displays the (log10) Euclidian distance between the $\argmax$ of \logei{} and its \mname{} variant. We note that, for small amounts ($\leq 256)$ of posterior samples, the error induced by RFF bias is relatively low, which is evidenced by all RFF variants being roughly equal in distance to the true acquisition function optimizer.


\subsection{Synthetic Functions with Known Priors}\label{sec:knownpi}
\begin{figure}[htbp]
  \centering
\includegraphics[width=\linewidth]{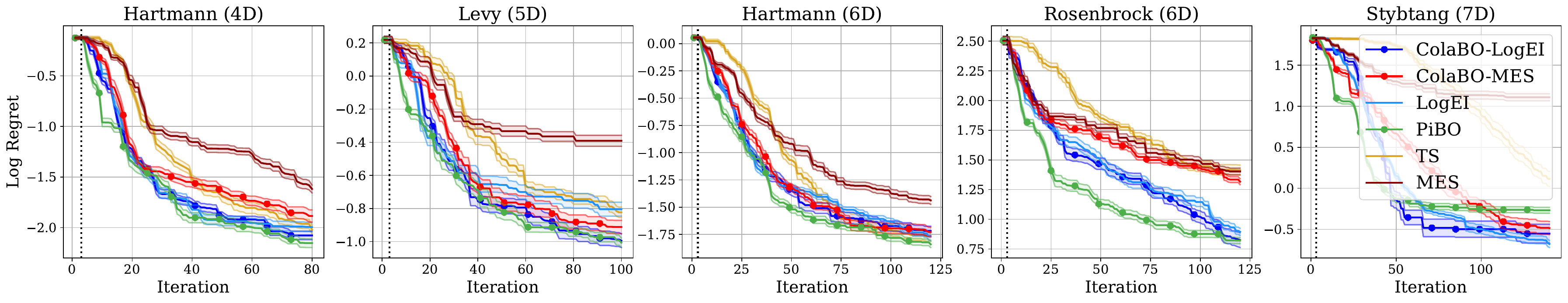} 
\caption{Performance on synthetic functions with well-located priors. Both \mname{}-\logei{} and \mname{}-\mes{} offer drastic speed-ups over their vanilla variants, and offer similar performance to \pibo{}. The ranking of \mname{} acquisition functions are generally consistent with their respective vanilla variants. This is most prominent on Rosenbrock (6D), where \mname{}-\mes{} struggles similarly to vanilla \mes{}.}
\label{fig:goodprior}
\end{figure}

We adapt a similar evaluation protocol to~\citet{hvarfner2022pibo}, and evaluate \mname{} for two types of user beliefs for synthetic tasks: well-located and poorly located priors over the optimal location, designed to emulate a well-informed and poorly-informed practitioner, respectively. The well-located prior is offset by a small (10\%) amount from the optimum, and the poorly located prior is maximally offset, while retaining its mode inside the search space. Complete details on the priors can be found in Appendix~\ref{app:priors}.
On well-located priors, both \mname{}-\logei{} and \mname{}-\mes{} demonstrate substantially improved performance relative to their vanilla counterparts, comparable to \pibo{} on all benchmarks. On poorly located priors, \mname{} demonstrates superior robustness, recovering the performance of the vanilla acquisition function within the maximal budget of $20D$ iterations and clearly outperforming \pibo{}, which more frequently misled by the poor prior. In Appendix~\ref{app:maxvalue}, we also demonstrate \mname{} utilizing (accurate) beliefs over the optimal value: similarly to Figure~\ref{fig:goodprior}, \mname{} yields increased efficiency relative to baselines, albeit not as substantial. Moreover, we demonstrate its usage with batch evaluations on well-located priors in Sec.~\ref{app:batch}, showing that the drop in performance from batching evaluations is marginal at worst.
\begin{figure}[htbp]
  \centering
\includegraphics[width=\linewidth]{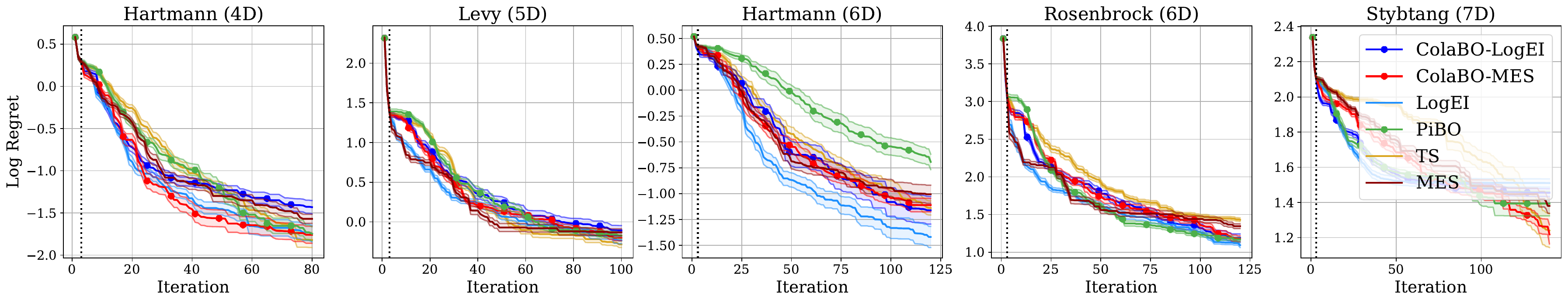} 
\caption{Performance on poorly located priors. \mname{} acquisition functions are more robust than \pibo{}, as it frequently recovers the performance of the vanilla acquisition function before the total budget is depleted. \mname{}-\logei{} struggles marginally on Hartmann (6D). \mname{}-\mes{} recovers the baseline on all tasks.}
\end{figure}
\subsection{Hyperparameter Tuning tasks}
For the real-world HPO tasks, we consider two different benchmarking suites: LCBench~\citep{Zimmer2020AutoPyTorchTM} and PD1~\citep{wang2023hyperbo}. For LCBench, we evaluate all methods on five deep learning tasks (6D). While the optima for these tasks are ultimately unknown, we utilize the priors provided in MF-Prior-Bench \footnote{\url{https://github.com/automl/mf-prior-bench}}~\citep{mallik2023priorband}, which are intended to provide a good starting point for further optimization. The chosen tasks were the five tasks with available priors of the best (good) strength, as per the benchmark suite. To emulate a realistic HPO setting, we consider a smaller optimization budget of $40$ iterations, and initialize all methods that utilize user beliefs with only one initial sample, that being the mode of the prior. 
\begin{figure}[htbp]
  \centering
\includegraphics[width=\linewidth]{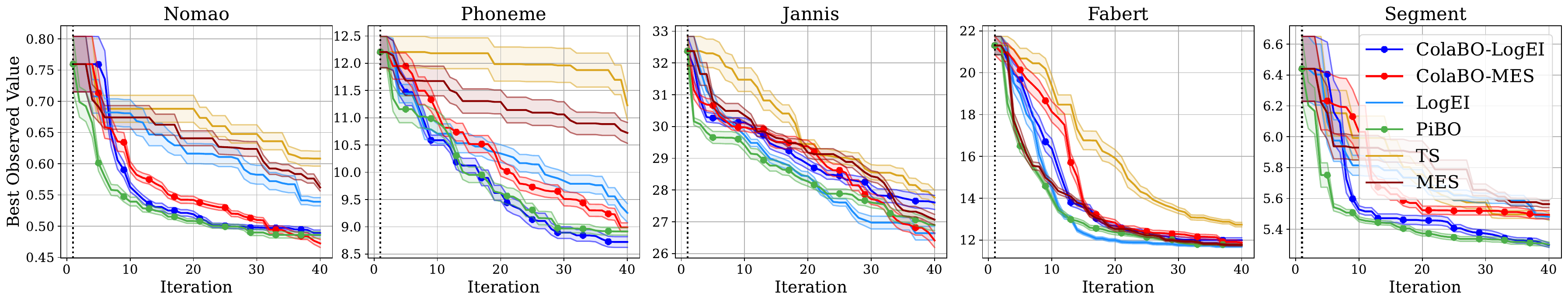} 
\caption{Performance on the 6D LCBench hyperparameter tuning tasks of various deep learning pipelines. \mname{} substantially improves on the non-prior baselines for 3 out of five tasks. \pibo{} performs best on aggregate, and achieves the best acceleration in performance at early iterations.}
\label{fig:lcbench}
\end{figure}
Figure~\ref{fig:lcbench} shows the performance of all methods on the LCBench tasks. \mname{} improves substantially on the baseline approaches for 3 out of 5 tasks. \pibo{} is the overall best-performing method, followed by \mname{}-\logei{}.

Lastly, we evaluate \mname{} on three $4D$ deep learning HPO tasks from the PD1~\citep{wang2023hyperbo} benchmarking suite, once again using priors from MF-Prior-Bench. The two \mname{} variants perform best in this evaluation, producing the best terminal performance on two tasks (CIFAR, LM1B), with all methods being tied on the third (CIFAR). \mname{} demonstrates consistent speed-ups compared to its vanilla counterparts, 
surpassing the terminal performance of the baseline within a third of the budget on CIFAR and LM1B.
\begin{figure}[tbp]
  \centering
\includegraphics[width=\linewidth]{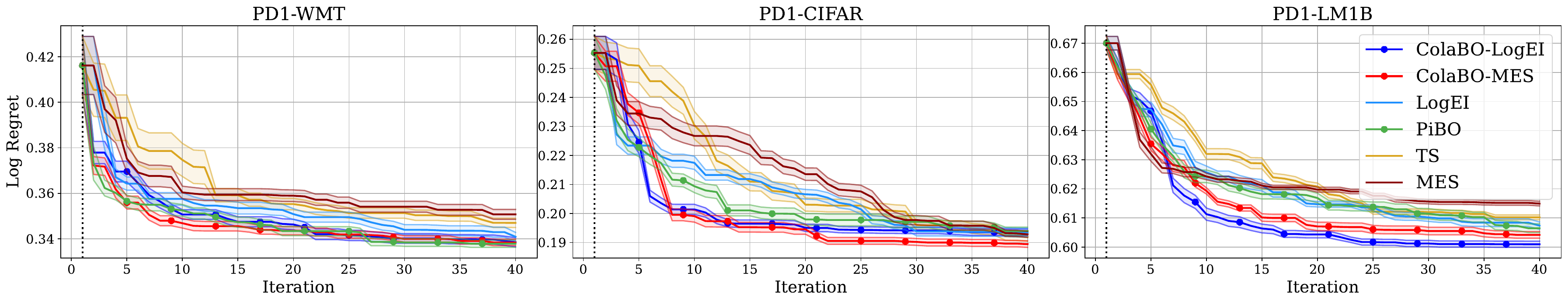} 
\caption{Performance on the 4D PD1 hyperparameter tuning tasks of various deep learning pipelines. \mname{} drastically accelerates optimization initially, finding configurations with close to terminal performance quickly. \pibo{} offers competitive performance, but lacks the rapid initial progress of \mname{} on CIFAR and LM1B.}
\vspace{-5mm}
\end{figure}
\section{Related Work}\label{related}
In BO, auxiliary prior information can be conveyed in multiple ways. We outline meta learning/transfer learning for BO based on data from previous experiments, and data-less approaches.

\paragraph{Learning from Previous Experiments}
Transfer learning and meta learning for BO aims to automatically extract and use knowledge from prior executions of BO by pre-training the model on data acquired from previous executions~\citep{swersky-nips13a,wistuba-pkdd15a,Perrone2019LearningSS,Feurer2015InitializingBH,feurer2018practical,rothfuss21pacoh,rothfuss2021fpacoh,wistuba2021fewshot,feurer2022practical}. Typically, meta- and transfer learning exploit relevant previous data for training the GP for the current task while retaining predictive uncertainty to account for imperfect task correlation. 

\paragraph{Expert Priors over Function Optimum}
Few previous works have proposed to inject explicit prior distributions over the location of an optimum into BO. In these cases, users explicitly define a prior that encodes their beliefs on where the optimum is more likely to be located. \citet{bergstra-nips11a} suggest an approach that supports prior beliefs from a fixed set of distributions, which affects the very initial stage of optimization. However, this approach cannot be combined with standard acquisition functions. BOPrO~\citep{souza2021bayesian} employs a similar structure that combines the user-provided prior distribution with a data-driven model into a pseudo-posterior. From the pseudo-posterior, configurations are selected using the EI acquisition function, using the formulation in \citet{bergstra-nips11a}. $\pi$BO~\citep{hvarfner2022pibo} suggests a general-purpose prior-weighted acquisition function, where the influence of the prior decreases over time. They provide convergence guarantees for when the framework is applied to the EI acquisition function. While effective, none of these approaches act on the surrogate model in a Bayesian-principled fashion, but strictly as heuristics. Moreover, they solely focus on priors over optimal inputs, thus offering less utility than \mname{}.
\paragraph{Priors over Optimal Value} Similarly  few works have addressed the issue of auxilliary knowledge of the optimal value. Both ~\citet{pmlr-v139-jeong21a} and \citet{pmlr-v119-nguyen20d} propose altering the GP and accompanying it with tailored acquisition functions. \citet{pmlr-v139-jeong21a} employ variational inference, proposing distinct variational families depending on the type of knowledge pertaining to the optimal value. \citet{pmlr-v119-nguyen20d} use a parabolic transformation of the output space to ensure an upper bound is preserved. Unlike \mname{}, neither of these methods is general enough to accompany arbitrary user priors to guide the optimization.

\section{Conclusion, Limitations and Future Work}
We presented \mname{}, a flexible BO framework that allows practitioners to inject beliefs over function properties in a Bayesian-principled manner, allowing for increased efficiency in the BO procedure. \mname{} works across a collection of MC acquisition functions, inheriting their flexibility in batch optimization and ability to work with non-Gaussian posteriors. It demonstrates competitive performance for well-located priors, using them to substantially accelerate optimization. Moreover, it retains approximately baseline performance when applied to detrimental priors, demonstrating greater robustness than \pibo{}. \mname{} crucially relies on multiple steps of MC. While flexible, this approach drives computational expense in order to assert sufficient accuracy, requiring tens of seconds per evaluation to achieve desired accuracy, depending on the size of the benchmark. Moreover, obtaining draws from $\rho_{\bm{x}}^*$ scales exponentially in the dimensionality of the prior. While practitioners are unlikely to specify priors over more than a handful of variables, \mname{} may become impractical when priors of higher dimensionality are employed. 
Paths for future work could involve more accurate and efficient sampling procedures~\citep{lin2023sampling} from the belief-weighted prior, as well as variational~\citep{pmlr-v5-titsias09a} or pre-trained~\cite{muller2022transformers, pmlr-v202-muller23a} approaches to obtain a representative belief-biased model with an analytical posterior. This would likely bring down the runtime of \mname{} and broaden its potential use. Lastly, applying \mname{} to multi-fidelity optimization~\citep{kandasamy-nips16a, mallik2023priorband} offers an additional avenue for increased efficiency which would further increase its viability on costly deep learning pipelines.


\bibliography{bibliography/local,bibliography/lib,bibliography/proc,bibliography/strings}

\begin{thebibliography}{76}
\providecommand{\natexlab}[1]{#1}
\providecommand{\url}[1]{\texttt{#1}}
\expandafter\ifx\csname urlstyle\endcsname\relax
  \providecommand{\doi}[1]{doi: #1}\else
  \providecommand{\doi}{doi: \begingroup \urlstyle{rm}\Url}\fi

\bibitem[Ament et~al.(2023)Ament, Daulton, Eriksson, Balandat, and Bakshy]{anonymous2023unexpected}
Sebastian Ament, Samuel Daulton, David Eriksson, Maximilian Balandat, and Eytan Bakshy.
\newblock Unexpected improvements to expected improvement for bayesian optimization.
\newblock In \emph{Thirty-seventh Conference on Neural Information Processing Systems}, 2023.
\newblock URL \url{https://openreview.net/forum?id=1vyAG6j9PE}.

\bibitem[Astudillo \& Frazier(2021)Astudillo and Frazier]{astudillo2021bayesian}
Raul Astudillo and Peter Frazier.
\newblock Bayesian optimization of function networks.
\newblock \emph{Advances in neural information processing systems}, 34:\penalty0 14463--14475, 2021.

\bibitem[Balandat et~al.(2020)Balandat, Karrer, Jiang, Daulton, Letham, Wilson, and Bakshy]{balandat2020botorch}
M.~Balandat, B.~Karrer, D.~R. Jiang, S.~Daulton, B.~Letham, A.~G. Wilson, and E.~Bakshy.
\newblock Botorch: A framework for efficient monte-carlo bayesian optimization.
\newblock In \emph{Advances in Neural Information Processing Systems}, 2020.
\newblock URL \url{http://arxiv.org/abs/1910.06403}.

\bibitem[Bergstra et~al.(2011{\natexlab{a}})Bergstra, Bardenet, Bengio, and K{\'e}gl]{bergstra-nips11a}
J.~Bergstra, R.~Bardenet, Y.~Bengio, and B.~K{\'e}gl.
\newblock Algorithms for hyper-parameter optimization.
\newblock In J.~Shawe-Taylor, R.~Zemel, P.~Bartlett, F.~Pereira, and K.~Weinberger (eds.), \emph{Proceedings of the 25th International Conference on Advances in Neural Information Processing Systems ({N}eur{IPS}'11)}, pp.\  2546--2554, 2011{\natexlab{a}}.

\bibitem[Bergstra et~al.(2011{\natexlab{b}})Bergstra, Bardenet, Bengio, and K\'{e}gl]{NIPS2011_86e8f7ab}
James Bergstra, R\'{e}mi Bardenet, Yoshua Bengio, and Bal\'{a}zs K\'{e}gl.
\newblock {Algorithms for Hyper-Parameter Optimization}.
\newblock In \emph{Advances in Neural Information Processing Systems (NeurIPS)}, volume~24. Curran Associates, Inc., 2011{\natexlab{b}}.

\bibitem[Brochu et~al.(2010)Brochu, Cora, and de~Freitas]{brochu-arXiv10a}
E.~Brochu, V.~Cora, and N.~de~Freitas.
\newblock A tutorial on {Bayesian} optimization of expensive cost functions, with application to active user modeling and hierarchical reinforcement learning.
\newblock \emph{arXiv:1012.2599v1 {[cs.LG]}}, 2010.

\bibitem[Bull(2011)]{bull-jmlr11a}
Adam~D. Bull.
\newblock Convergence rates of efficient global optimization algorithms.
\newblock 12:\penalty0 2879--2904, 2011.

\bibitem[Calandra et~al.(2014)Calandra, Gopalan, Seyfarth, Peters, and Deisenroth]{calandra-lion14a}
R.~Calandra, N.~Gopalan, A.~Seyfarth, J.~Peters, and M.~Deisenroth.
\newblock {B}ayesian gait optimization for bipedal locomotion.
\newblock In P.~Pardalos and M.~Resende (eds.), \emph{Proceedings of the Eighth International Conference on Learning and Intelligent Optimization ({LION}'14)}, 2014.

\bibitem[Ejjeh et~al.(2022)Ejjeh, Medvinsky, Councilman, Nehra, Sharma, Adve, Nardi, Nurvitadhi, and Rutenbar]{ejjeh2022hpvm2fpga}
Adel Ejjeh, Leon Medvinsky, Aaron Councilman, Hemang Nehra, Suraj Sharma, Vikram Adve, Luigi Nardi, Eriko Nurvitadhi, and Rob~A Rutenbar.
\newblock Hpvm2fpga: Enabling true hardware-agnostic fpga programming.
\newblock In \emph{Proceedings of the 33rd IEEE International Conference on Application-specific Systems, Architectures, and Processors}, 2022.

\bibitem[Eriksson et~al.(2019)Eriksson, Pearce, Gardner, Turner, and Poloczek]{eriksson2019scalable}
David Eriksson, Michael Pearce, Jacob Gardner, Ryan~D Turner, and Matthias Poloczek.
\newblock Scalable global optimization via local {Bayesian} optimization.
\newblock In \emph{Advances in Neural Information Processing Systems}, pp.\  5496--5507, 2019.
\newblock URL \url{http://papers.nips.cc/paper/8788-scalable-global-optimization-via-local-bayesian-optimization.pdf}.

\bibitem[Feurer et~al.(2015)Feurer, Springenberg, and Hutter]{Feurer2015InitializingBH}
M.~Feurer, Jost~Tobias Springenberg, and F.~Hutter.
\newblock Initializing bayesian hyperparameter optimization via meta-learning.
\newblock In \emph{Proceedings of the Twenty-Ninth {AAAI} Conference on Artificial Intelligence}, pp.\  1128--1135, 2015.

\bibitem[Feurer et~al.(2018)Feurer, Letham, Hutter, and Bakshy]{feurer2018practical}
M.~Feurer, B.~Letham, F.~Hutter, and E.~Bakshy.
\newblock Practical transfer learning for bayesian optimization.
\newblock \emph{ArXiv abs/1802.02219}, 2018.

\bibitem[Feurer et~al.(2022)Feurer, Letham, Hutter, and Bakshy]{feurer2022practical}
Matthias Feurer, Benjamin Letham, Frank Hutter, and Eytan Bakshy.
\newblock Practical transfer learning for {B}ayesian optimization.
\newblock \emph{arXiv preprint 1802.02219}, 2022.

\bibitem[Frazier et~al.(2009)Frazier, Powell, and Dayanik]{frazier2009knowledge}
Peter Frazier, Warren Powell, and Savas Dayanik.
\newblock The knowledge-gradient policy for correlated normal beliefs.
\newblock \emph{INFORMS journal on Computing}, 21\penalty0 (4):\penalty0 599--613, 2009.

\bibitem[Garnett(2022)]{garnett-book22a}
R.~Garnett.
\newblock \emph{{Bayesian Optimization}}.
\newblock Cambridge University Press, 2022.
\newblock Available for free at \url{https://bayesoptbook.com/}.

\bibitem[Griffiths \& Hern{\'a}ndez-Lobato(2020)Griffiths and Hern{\'a}ndez-Lobato]{griffiths2020constrained}
Ryan-Rhys Griffiths and Jos{\'e}~Miguel Hern{\'a}ndez-Lobato.
\newblock Constrained bayesian optimization for automatic chemical design using variational autoencoders.
\newblock \emph{Chemical Science}, 2020.

\bibitem[Hennig \& Schuler(2012)Hennig and Schuler]{entropysearch}
P.~Hennig and C.~J. Schuler.
\newblock Entropy search for information-efficient global optimization.
\newblock \emph{Journal of Machine Learning Research}, 13\penalty0 (1):\penalty0 1809–1837, June 2012.
\newblock ISSN 1532-4435.

\bibitem[Hern\'{a}ndez-Lobato et~al.(2014)Hern\'{a}ndez-Lobato, Hoffman, and Ghahramani]{pes}
J.~M. Hern\'{a}ndez-Lobato, M.~W. Hoffman, and Z.~Ghahramani.
\newblock Predictive entropy search for efficient global optimization of black-box functions.
\newblock In \emph{Advances in Neural Information Processing Systems}, 2014.
\newblock URL \url{https://proceedings.neurips.cc/paper/2014/file/069d3bb002acd8d7dd095917f9efe4cb-Paper.pdf}.

\bibitem[Hern{\'a}ndez-Lobato et~al.(2015)Hern{\'a}ndez-Lobato, Gelbart, Hoffman, Adams, and Ghahramani]{hernandez2015predictive}
Jos{\'e}~Miguel Hern{\'a}ndez-Lobato, Michael Gelbart, Matthew Hoffman, Ryan Adams, and Zoubin Ghahramani.
\newblock Predictive entropy search for bayesian optimization with unknown constraints.
\newblock In \emph{International conference on machine learning}, pp.\  1699--1707. PMLR, 2015.

\bibitem[Huang et~al.(2022)Huang, Filstroff, Mikkola, Zheng, and Kaski]{huang2022bayesian}
Daolang Huang, Louis Filstroff, Petrus Mikkola, Runkai Zheng, and Samuel Kaski.
\newblock Bayesian optimization augmented with actively elicited expert knowledge, 2022.

\bibitem[Hutter et~al.(2011)Hutter, Hoos, and Leyton-Brown]{hutter-lion11a}
F.~Hutter, H.~Hoos, and K.~Leyton-Brown.
\newblock Sequential model-based optimization for general algorithm configuration.
\newblock In C.~Coello (ed.), \emph{Proceedings of the Fifth International Conference on Learning and Intelligent Optimization ({LION}'11)}, volume 6683, pp.\  507--523, 2011.

\bibitem[Hvarfner et~al.(2022{\natexlab{a}})Hvarfner, Hutter, and Nardi]{hvarfner2022joint}
Carl Hvarfner, Frank Hutter, and Luigi Nardi.
\newblock Joint entropy search for maximally-informed bayesian optimization.
\newblock In \emph{Proceedings of the 36th International Conference on Neural Information Processing Systems}, 2022{\natexlab{a}}.

\bibitem[Hvarfner et~al.(2022{\natexlab{b}})Hvarfner, Stoll, Souza, Lindauer, Hutter, and Nardi]{hvarfner2022pibo}
Carl Hvarfner, Danny Stoll, Artur Souza, Marius Lindauer, Frank Hutter, and Luigi Nardi.
\newblock {{PiBO}: Augmenting Acquisition Functions with User Beliefs for Bayesian Optimization}.
\newblock In \emph{International Conference on Learning Representations}, 2022{\natexlab{b}}.

\bibitem[Hvarfner et~al.(2023)Hvarfner, Hellsten, Hutter, and Nardi]{hvarfner2023selfcorrecting}
Carl Hvarfner, Erik Hellsten, Frank Hutter, and Luigi Nardi.
\newblock Self-correcting bayesian optimization through bayesian active learning.
\newblock In \emph{Thirty-seventh Conference on Neural Information Processing Systems}, 2023.
\newblock URL \url{https://openreview.net/forum?id=dX9MjUtP1A}.

\bibitem[Jeong \& Kim(2021)Jeong and Kim]{pmlr-v139-jeong21a}
Taewon Jeong and Heeyoung Kim.
\newblock Objective bound conditional gaussian process for bayesian optimization.
\newblock In Marina Meila and Tong Zhang (eds.), \emph{Proceedings of the 38th International Conference on Machine Learning}, volume 139 of \emph{Proceedings of Machine Learning Research}, pp.\  4819--4828. PMLR, 18--24 Jul 2021.
\newblock URL \url{https://proceedings.mlr.press/v139/jeong21a.html}.

\bibitem[Jones et~al.(1998)Jones, Schonlau, and Welch]{jonesei}
D.~Jones, M.~Schonlau, and W.~Welch.
\newblock Efficient global optimization of expensive black-box functions.
\newblock \emph{Journal of Global Optimization}, 13:\penalty0 455--492, 12 1998.
\newblock \doi{10.1023/A:1008306431147}.

\bibitem[Journel \& Huijbregts(1976)Journel and Huijbregts]{etde_5214736}
A~G Journel and C~J Huijbregts.
\newblock Mining geostatistics, Jan 1976.

\bibitem[Kandasamy et~al.(2016)Kandasamy, Dasarathy, Oliva, Schneider, and Póczos]{kandasamy-nips16a}
K.~Kandasamy, G.~Dasarathy, J.~Oliva, J.~Schneider, and B.~Póczos.
\newblock Gaussian {Process} {Bandit} {Optimisation} with {Multi}-fidelity {Evaluations}.
\newblock In D.~Lee, M.~Sugiyama, U.~von Luxburg, I.~Guyon, and R.~Garnett (eds.), \emph{Proceedings of the 30th International Conference on Advances in Neural Information Processing Systems ({N}eur{IPS}'16)}, pp.\  992--1000, 2016.

\bibitem[Kandasamy et~al.(2018)Kandasamy, Krishnamurthy, Schneider, and Póczos]{kandasamy-aistats18a}
K.~Kandasamy, A.~Krishnamurthy, J.~Schneider, and B.~Póczos.
\newblock Parallelised {B}ayesian optimisation via {T}hompson sampling.
\newblock In A.~Storkey and F~Perez-Cruz (eds.), \emph{Proceedings of the 21st International Conference on Artificial Intelligence and Statistics ({AISTATS})}, volume~84, pp.\  133--142. Proceedings of Machine Learning Research, 2018.

\bibitem[Kingma \& Welling(2014)Kingma and Welling]{vae}
Diederik~P Kingma and Max Welling.
\newblock Auto-encoding variational bayes, 2014.
\newblock URL \url{https://arxiv.org/abs/1312.6114}.

\bibitem[Kumar et~al.(2022)Kumar, Rana, Shilton, and Venkatesh]{NEURIPS2022_6751611b}
Arun Kumar, Santu Rana, Alistair Shilton, and Svetha Venkatesh.
\newblock Human-ai collaborative bayesian optimisation.
\newblock In S.~Koyejo, S.~Mohamed, A.~Agarwal, D.~Belgrave, K.~Cho, and A.~Oh (eds.), \emph{Advances in Neural Information Processing Systems}, volume~35, pp.\  16233--16245. Curran Associates, Inc., 2022.
\newblock URL \url{https://proceedings.neurips.cc/paper_files/paper/2022/file/6751611b394a3464cea53eed91cf163c-Paper-Conference.pdf}.

\bibitem[Kushner(1964)]{pi}
H.~J. Kushner.
\newblock {A New Method of Locating the Maximum Point of an Arbitrary Multipeak Curve in the Presence of Noise}.
\newblock \emph{Journal of Basic Engineering}, 86\penalty0 (1):\penalty0 97--106, 03 1964.
\newblock ISSN 0021-9223.
\newblock \doi{10.1115/1.3653121}.
\newblock URL \url{https://doi.org/10.1115/1.3653121}.

\bibitem[Letham et~al.(2018)Letham, Brian, Ottoni, and Bakshy]{letham-ba18a}
B.~Letham, K.~Brian, G.~Ottoni, and E.~Bakshy.
\newblock Constrained {B}ayesian optimization with noisy experiments.
\newblock \emph{Bayesian {A}nalysis}, 2018.

\bibitem[Lin et~al.(2023)Lin, Antorán, Padhy, Janz, Hernández-Lobato, and Terenin]{lin2023sampling}
Jihao~Andreas Lin, Javier Antorán, Shreyas Padhy, David Janz, José~Miguel Hernández-Lobato, and Alexander Terenin.
\newblock Sampling from gaussian process posteriors using stochastic gradient descent.
\newblock In \emph{Thirty-seventh Conference on Neural Information Processing Systems}, 2023.
\newblock URL \url{https://openreview.net/forum?id=Sf9goJtTCE}.

\bibitem[Lindauer et~al.(2022)Lindauer, Eggensperger, Feurer, Biedenkapp, Deng, Benjamins, Ruhkopf, Sass, and Hutter]{lindauer_smac3}
Marius Lindauer, Katharina Eggensperger, Matthias Feurer, André Biedenkapp, Difan Deng, Carolin Benjamins, Tim Ruhkopf, René Sass, and Frank Hutter.
\newblock Smac3: A versatile bayesian optimization package for hyperparameter optimization.
\newblock \emph{Journal of Machine Learning Research}, 23\penalty0 (54):\penalty0 1--9, 2022.
\newblock URL \url{http://jmlr.org/papers/v23/21-0888.html}.

\bibitem[Mallik et~al.(2023)Mallik, Bergman, Hvarfner, Stoll, Janowski, Lindauer, Nardi, and Hutter]{mallik2023priorband}
Neeratyoy Mallik, Edward Bergman, Carl Hvarfner, Danny Stoll, Maciej Janowski, Marius Lindauer, Luigi Nardi, and Frank Hutter.
\newblock Priorband: Practical hyperparameter optimization in the age of deep learning.
\newblock \emph{arXiv preprint 2306.12370}, 2023.

\bibitem[Mat{\'e}rn(1960)]{matern1960spatial}
B.~Mat{\'e}rn.
\newblock Spatial variation.
\newblock \emph{Meddelanden fran Statens Skogsforskningsinstitut}, 1960.

\bibitem[Mayr et~al.(2022)Mayr, Hvarfner, Chatzilygeroudis, Nardi, and Krueger]{mayr2022learning}
Matthias Mayr, Carl Hvarfner, Konstantinos Chatzilygeroudis, Luigi Nardi, and Volker Krueger.
\newblock Learning skill-based industrial robot tasks with user priors.
\newblock \emph{IEEE 18th International Conference on Automation Science and Engineering}, 2022.
\newblock URL \url{https://arxiv.org/abs/2208.01605}.

\bibitem[Mockus et~al.(1978)Mockus, Tiesis, and Zilinskas]{Mockus1978}
J.~Mockus, V.~Tiesis, and A.~Zilinskas.
\newblock The application of {B}ayesian methods for seeking the extremum.
\newblock \emph{Towards Global Optimization}, 2\penalty0 (117-129):\penalty0 2, 1978.

\bibitem[Moss et~al.(2021)Moss, Leslie, Gonzalez, and Rayson]{moss2021gibbon}
Henry~B. Moss, David~S. Leslie, Javier Gonzalez, and Paul Rayson.
\newblock Gibbon: General-purpose information-based bayesian optimisation.
\newblock \emph{Journal of Machine Learning Research}, 22\penalty0 (235):\penalty0 1--49, 2021.
\newblock URL \url{http://jmlr.org/papers/v22/21-0120.html}.

\bibitem[M{\"u}ller et~al.(2022)M{\"u}ller, Hollmann, Arango, Grabocka, and Hutter]{muller2022transformers}
Samuel M{\"u}ller, Noah Hollmann, Sebastian~Pineda Arango, Josif Grabocka, and Frank Hutter.
\newblock Transformers can do bayesian inference.
\newblock In \emph{International Conference on Learning Representations}, 2022.
\newblock URL \url{https://openreview.net/forum?id=KSugKcbNf9}.

\bibitem[M\"{u}ller et~al.(2023)M\"{u}ller, Feurer, Hollmann, and Hutter]{pmlr-v202-muller23a}
Samuel M\"{u}ller, Matthias Feurer, Noah Hollmann, and Frank Hutter.
\newblock {PFN}s4{BO}: In-context learning for {B}ayesian optimization.
\newblock In Andreas Krause, Emma Brunskill, Kyunghyun Cho, Barbara Engelhardt, Sivan Sabato, and Jonathan Scarlett (eds.), \emph{Proceedings of the 40th International Conference on Machine Learning}, volume 202 of \emph{Proceedings of Machine Learning Research}, pp.\  25444--25470. PMLR, 23--29 Jul 2023.
\newblock URL \url{https://proceedings.mlr.press/v202/muller23a.html}.

\bibitem[Mutny \& Krause(2018)Mutny and Krause]{mutny18efficient}
Mojmir Mutny and Andreas Krause.
\newblock Efficient high dimensional bayesian optimization with additivity and quadrature fourier features.
\newblock In S.~Bengio, H.~Wallach, H.~Larochelle, K.~Grauman, N.~Cesa-Bianchi, and R.~Garnett (eds.), \emph{Advances in Neural Information Processing Systems}, volume~31. Curran Associates, Inc., 2018.
\newblock URL \url{https://proceedings.neurips.cc/paper_files/paper/2018/file/4e5046fc8d6a97d18a5f54beaed54dea-Paper.pdf}.

\bibitem[Nardi et~al.(2019)Nardi, Koeplinger, and Olukotun]{nardi2019practical}
L.~Nardi, D.~Koeplinger, and K.~Olukotun.
\newblock Practical design space exploration.
\newblock In \emph{2019 IEEE 27th International Symposium on Modeling, Analysis, and Simulation of Computer and Telecommunication Systems (MASCOTS)}, pp.\  347--358. IEEE, 2019.

\bibitem[Neiswanger et~al.(2021)Neiswanger, Wang, and Ermon]{pmlr-v139-neiswanger21a}
Willie Neiswanger, Ke~Alexander Wang, and Stefano Ermon.
\newblock Bayesian algorithm execution: Estimating computable properties of black-box functions using mutual information.
\newblock In Marina Meila and Tong Zhang (eds.), \emph{Proceedings of the 38th International Conference on Machine Learning}, volume 139 of \emph{Proceedings of Machine Learning Research}, pp.\  8005--8015. PMLR, 18--24 Jul 2021.
\newblock URL \url{https://proceedings.mlr.press/v139/neiswanger21a.html}.

\bibitem[Nguyen \& Osborne(2020)Nguyen and Osborne]{pmlr-v119-nguyen20d}
Vu~Nguyen and Michael~A. Osborne.
\newblock Knowing the what but not the where in {B}ayesian optimization.
\newblock In Hal~Daumé III and Aarti Singh (eds.), \emph{Proceedings of the 37th International Conference on Machine Learning}, volume 119 of \emph{Proceedings of Machine Learning Research}, pp.\  7317--7326. PMLR, 13--18 Jul 2020.
\newblock URL \url{https://proceedings.mlr.press/v119/nguyen20d.html}.

\bibitem[Oh et~al.(2018)Oh, Gavves, and Welling]{oh2018bock}
C.~Oh, E.~Gavves, and M.~Welling.
\newblock {BOCK} : Bayesian optimization with cylindrical kernels.
\newblock In \emph{International Conference on Machine Learning}, pp.\  3865--3874, 2018.

\bibitem[Perrone et~al.(2019)Perrone, Shen, Seeger, Archambeau, and Jenatton]{Perrone2019LearningSS}
V.~Perrone, H.~Shen, M.~Seeger, C.~Archambeau, and R.~Jenatton.
\newblock Learning search spaces for bayesian optimization: Another view of hyperparameter transfer learning.
\newblock In \emph{Advances in Neural Information Processing Systems}, 2019.

\bibitem[Rahimi \& Recht(2007)Rahimi and Recht]{rahimi2007rff}
Ali Rahimi and Benjamin Recht.
\newblock Random features for large-scale kernel machines.
\newblock In J.~Platt, D.~Koller, Y.~Singer, and S.~Roweis (eds.), \emph{Advances in Neural Information Processing Systems}, volume~20. Curran Associates, Inc., 2007.
\newblock URL \url{https://proceedings.neurips.cc/paper_files/paper/2007/file/013a006f03dbc5392effeb8f18fda755-Paper.pdf}.

\bibitem[Rasmussen \& Williams(2006)Rasmussen and Williams]{rasmussen-book06a}
C.~Rasmussen and C.~Williams.
\newblock \emph{Gaussian Processes for Machine Learning}.
\newblock The MIT Press, 2006.

\bibitem[Rezende et~al.(2014)Rezende, Mohamed, and Wierstra]{pmlr-v32-rezende14}
Danilo~Jimenez Rezende, Shakir Mohamed, and Daan Wierstra.
\newblock Stochastic backpropagation and approximate inference in deep generative models.
\newblock In Eric~P. Xing and Tony Jebara (eds.), \emph{Proceedings of the 31st International Conference on Machine Learning}, volume~32 of \emph{Proceedings of Machine Learning Research}, pp.\  1278--1286, Bejing, China, 22--24 Jun 2014. PMLR.
\newblock URL \url{https://proceedings.mlr.press/v32/rezende14.html}.

\bibitem[Rothfuss et~al.(2021{\natexlab{a}})Rothfuss, Fortuin, Josifoski, and Krause]{rothfuss21pacoh}
Jonas Rothfuss, Vincent Fortuin, Martin Josifoski, and Andreas Krause.
\newblock Pacoh: Bayes-optimal meta-learning with pac-guarantees.
\newblock In \emph{Proceedings of the 38th International Conference on Machine Learning}, pp.\  9116--9126, 2021{\natexlab{a}}.

\bibitem[Rothfuss et~al.(2021{\natexlab{b}})Rothfuss, Heyn, Chen, and Krause]{rothfuss2021fpacoh}
Jonas Rothfuss, Dominique Heyn, Jinfan Chen, and Andreas Krause.
\newblock Meta-learning reliable priors in the function space.
\newblock In \emph{Advances in Neural Information Processing Systems}, volume~34, 2021{\natexlab{b}}.

\bibitem[Ru et~al.(2021)Ru, Wan, Dong, and Osborne]{ru2021interpretable}
Binxin Ru, Xingchen Wan, Xiaowen Dong, and Michael Osborne.
\newblock Interpretable neural architecture search via bayesian optimisation with weisfeiler-lehman kernels.
\newblock In \emph{International Conference on Learning Representations}, 2021.
\newblock URL \url{https://openreview.net/forum?id=j9Rv7qdXjd}.

\bibitem[Shahriari et~al.(2016)Shahriari, Swersky, Wang, Adams, and de~Freitas]{shahriari-ieee16a}
B.~Shahriari, K.~Swersky, Z.~Wang, R.~Adams, and N.~de~Freitas.
\newblock Taking the human out of the loop: {A} review of {B}ayesian optimization.
\newblock \emph{Proceedings of the {IEEE}}, 104\penalty0 (1):\penalty0 148--175, 2016.

\bibitem[Smith(2018)]{smith2018disciplined}
L.~Smith.
\newblock A disciplined approach to neural network hyper-parameters: Part 1--learning rate, batch size, momentum, and weight decay.
\newblock \emph{arXiv preprint arXiv:1803.09820}, 2018.

\bibitem[Snoek et~al.(2012)Snoek, Larochelle, and Adams]{snoek-nips12a}
J.~Snoek, H.~Larochelle, and R.~Adams.
\newblock Practical {B}ayesian optimization of machine learning algorithms.
\newblock In P.~Bartlett, F.~Pereira, C.~Burges, L.~Bottou, and K.~Weinberger (eds.), \emph{Proceedings of the 26th International Conference on Advances in Neural Information Processing Systems ({N}eur{IPS}'12)}, pp.\  2960--2968, 2012.

\bibitem[Souza et~al.(2021)Souza, Nardi, Oliveira, Olukotun, Lindauer, and Hutter]{souza2021bayesian}
A.~Souza, L.~Nardi, L.~Oliveira, K.~Olukotun, M.~Lindauer, and F.~Hutter.
\newblock Bayesian optimization with a prior for the optimum.
\newblock In \emph{Machine Learning and Knowledge Discovery in Databases. Research Track - European Conference, {ECML} {PKDD} 2021, Bilbao, Spain, September 13-17, 2021, Proceedings, Part {III}}, volume 12977 of \emph{Lecture Notes in Computer Science}, pp.\  265--296. Springer, 2021.

\bibitem[Srinivas et~al.(2012)Srinivas, Krause, Kakade, and Seeger]{Srinivas_2012}
N.~Srinivas, A.~Krause, S.~M. Kakade, and M.~W. Seeger.
\newblock Information-theoretic regret bounds for gaussian process optimization in the bandit setting.
\newblock \emph{IEEE Transactions on Information Theory}, 58\penalty0 (5):\penalty0 3250–3265, May 2012.
\newblock ISSN 1557-9654.
\newblock \doi{10.1109/tit.2011.2182033}.
\newblock URL \url{http://dx.doi.org/10.1109/TIT.2011.2182033}.

\bibitem[Swersky et~al.(2013)Swersky, Snoek, and Adams]{swersky-nips13a}
K.~Swersky, J.~Snoek, and R.~Adams.
\newblock Multi-task {Bayesian} optimization.
\newblock In C.~Burges, L.~Bottou, M.~Welling, Z.~Ghahramani, and K.~Weinberger (eds.), \emph{Proceedings of the 27th International Conference on Advances in Neural Information Processing Systems ({N}eur{IPS}'13)}, pp.\  2004--2012, 2013.

\bibitem[Takeno et~al.(2020)Takeno, Fukuoka, Tsukada, Koyama, Shiga, Takeuchi, and Karasuyama]{pmlr-v119-takeno20a}
Shion Takeno, Hitoshi Fukuoka, Yuhki Tsukada, Toshiyuki Koyama, Motoki Shiga, Ichiro Takeuchi, and Masayuki Karasuyama.
\newblock Multi-fidelity {B}ayesian optimization with max-value entropy search and its parallelization.
\newblock In Hal~Daumé III and Aarti Singh (eds.), \emph{Proceedings of the 37th International Conference on Machine Learning}, volume 119 of \emph{Proceedings of Machine Learning Research}, pp.\  9334--9345. PMLR, 13--18 Jul 2020.
\newblock URL \url{https://proceedings.mlr.press/v119/takeno20a.html}.

\bibitem[Takeno et~al.(2022)Takeno, Tamura, Shitara, and Karasuyama]{pmlr-v162-takeno22a}
Shion Takeno, Tomoyuki Tamura, Kazuki Shitara, and Masayuki Karasuyama.
\newblock Sequential and parallel constrained max-value entropy search via information lower bound.
\newblock In Kamalika Chaudhuri, Stefanie Jegelka, Le~Song, Csaba Szepesvari, Gang Niu, and Sivan Sabato (eds.), \emph{Proceedings of the 39th International Conference on Machine Learning}, volume 162 of \emph{Proceedings of Machine Learning Research}, pp.\  20960--20986. PMLR, 17--23 Jul 2022.
\newblock URL \url{https://proceedings.mlr.press/v162/takeno22a.html}.

\bibitem[Thompson(1933)]{thompson1933likelihood}
W.~Thompson.
\newblock On the likelihood that one unknown probability exceeds another in view of the evidence of two samples.
\newblock \emph{Biometrika}, 25\penalty0 (3/4):\penalty0 285--294, 1933.

\bibitem[Titsias(2009)]{pmlr-v5-titsias09a}
Michalis Titsias.
\newblock Variational learning of inducing variables in sparse gaussian processes.
\newblock In David van Dyk and Max Welling (eds.), \emph{Proceedings of the Twelth International Conference on Artificial Intelligence and Statistics}, volume~5 of \emph{Proceedings of Machine Learning Research}, pp.\  567--574, Hilton Clearwater Beach Resort, Clearwater Beach, Florida USA, 16--18 Apr 2009. PMLR.
\newblock URL \url{https://proceedings.mlr.press/v5/titsias09a.html}.

\bibitem[Tu et~al.(2022)Tu, Gandy, Kantas, and Shafei]{tu2022joint}
Ben Tu, Axel Gandy, Nikolas Kantas, and Behrang Shafei.
\newblock Joint entropy search for multi-objective bayesian optimization.
\newblock In Alice~H. Oh, Alekh Agarwal, Danielle Belgrave, and Kyunghyun Cho (eds.), \emph{Advances in Neural Information Processing Systems}, 2022.
\newblock URL \url{https://openreview.net/forum?id=ZChgD8OoGds}.

\bibitem[Wang et~al.(2019)Wang, Ming, Jin, Shen, Liu, Smith, Veeramachaneni, and Qu]{atmseer}
Q.~Wang, Y.~Ming, Z.~Jin, Q.~Shen, D.~Liu, M.~J. Smith, K.~Veeramachaneni, and H.~Qu.
\newblock Atmseer: Increasing transparency and controllability in automated machine learning.
\newblock In \emph{Proceedings of the 2019 CHI Conference on Human Factors in Computing Systems}, CHI '19, pp.\  1–12. Association for Computing Machinery, 2019.

\bibitem[Wang \& Jegelka(2017)Wang and Jegelka]{wang2017maxvalue}
Zi~Wang and Stefanie Jegelka.
\newblock Max-value entropy search for efficient bayesian optimization.
\newblock In \emph{International Conference on Machine Learning (ICML)}, 2017.

\bibitem[Wang et~al.(2018)Wang, Gehring, Kohli, and Jegelka]{pmlr-v84-wang18c}
Zi~Wang, Clement Gehring, Pushmeet Kohli, and Stefanie Jegelka.
\newblock Batched large-scale bayesian optimization in high-dimensional spaces.
\newblock In Amos Storkey and Fernando Perez-Cruz (eds.), \emph{Proceedings of the Twenty-First International Conference on Artificial Intelligence and Statistics}, volume~84 of \emph{Proceedings of Machine Learning Research}, pp.\  745--754. PMLR, 09--11 Apr 2018.
\newblock URL \url{https://proceedings.mlr.press/v84/wang18c.html}.

\bibitem[Wang et~al.(2023)Wang, Dahl, Swersky, Lee, Nado, Gilmer, Snoek, and Ghahramani]{wang2023hyperbo}
Zi~Wang, George~E. Dahl, Kevin Swersky, Chansoo Lee, Zachary Nado, Justin Gilmer, Jasper Snoek, and Zoubin Ghahramani.
\newblock {Pre-trained Gaussian processes for Bayesian optimization}.
\newblock \emph{arXiv preprint arXiv:2109.08215}, 2023.

\bibitem[White et~al.(2021)White, Neiswanger, and Savani]{white-aaai21a}
C.~White, W.~Neiswanger, and Y.~Savani.
\newblock {BANANAS}: {Bayesian} optimization with neural architectures for neural architecture search.
\newblock In Q.~Yang, K.~Leyton-Brown, and Mausam (eds.), \emph{Proceedings of the Thirty-Fifth Conference on Artificial Intelligence ({AAAI}'21)}, pp.\  10293--10301. Association for the Advancement of Artificial Intelligence, {AAAI} Press, 2021.

\bibitem[Wilson et~al.(2018)Wilson, Hutter, and Deisenroth]{NEURIPS2018_498f2c21}
James Wilson, Frank Hutter, and Marc Deisenroth.
\newblock Maximizing acquisition functions for bayesian optimization.
\newblock In S.~Bengio, H.~Wallach, H.~Larochelle, K.~Grauman, N.~Cesa-Bianchi, and R.~Garnett (eds.), \emph{Advances in Neural Information Processing Systems}, volume~31. Curran Associates, Inc., 2018.
\newblock URL \url{https://proceedings.neurips.cc/paper/2018/file/498f2c21688f6451d9f5fd09d53edda7-Paper.pdf}.

\bibitem[Wilson et~al.(2017)Wilson, Moriconi, Hutter, and Deisenroth]{wilson2017reparam}
James~T. Wilson, Riccardo Moriconi, Frank Hutter, and Marc~Peter Deisenroth.
\newblock The reparameterization trick for acquisition functions, 2017.
\newblock URL \url{https://arxiv.org/abs/1712.00424}.

\bibitem[Wilson et~al.(2020)Wilson, Borovitskiy, Terenin, Mostowsky, and Deisenroth]{wilson2020efficiently}
James~T. Wilson, Viacheslav Borovitskiy, Alexander Terenin, Peter Mostowsky, and Marc~Peter Deisenroth.
\newblock Efficiently sampling functions from gaussian process posteriors.
\newblock In \emph{International Conference on Machine Learning}, 2020.
\newblock URL \url{https://arxiv.org/abs/2002.09309}.

\bibitem[Wistuba et~al.(2015)Wistuba, Schilling, and Schmidt{-}Thieme]{wistuba-pkdd15a}
M.~Wistuba, N.~Schilling, and L.~Schmidt{-}Thieme.
\newblock Hyperparameter search space pruning - {A} new component for sequential model-based hyperparameter optimization.
\newblock In A.~Appice, P.~Rodrigues, V.~Costa, J.~Gama, A.~Jorge, and C.~Soares (eds.), \emph{Machine Learning and Knowledge Discovery in Databases ({ECML}/{PKDD}'15)}, volume 9285, pp.\  104--119, 2015.

\bibitem[Wistuba \& Grabocka(2021)Wistuba and Grabocka]{wistuba2021fewshot}
Martin Wistuba and Josif Grabocka.
\newblock Few-shot bayesian optimization with deep kernel surrogates.
\newblock In \emph{International Conference on Learning Representations}, 2021.
\newblock URL \url{https://openreview.net/forum?id=bJxgv5C3sYc}.

\bibitem[Zimmer et~al.(2020)Zimmer, Lindauer, and Hutter]{Zimmer2020AutoPyTorchTM}
Lucas Zimmer, Marius~Thomas Lindauer, and Frank Hutter.
\newblock Auto-pytorch tabular: Multi-fidelity metalearning for efficient and robust autodl.
\newblock \emph{ArXiv}, abs/2006.13799, 2020.
\newblock URL \url{https://api.semanticscholar.org/CorpusID:220041844}.

\end{thebibliography}
\bibliographystyle{iclr2024_conference}

\appendix
\section{Experimental Setup}~\label{app:setup}
\subsection{Model}\label{app:model}
We outline the model used and the budget allocated to the various MC approximations involved with \mname{}. For all experiments, we utilize MAP estimation of the hyperparameters, and update the hyperparameters at every iteration of BO. All hyperparameters - lengthscale, outputscale and observation noise ($\hps = \{\bm{\ell}, \sigma_\varepsilon^2, \sigma_f^2\})$ are given conventional $\mathcal{LN}(0, 1)$ prior, applied on normalized inputs and standardized outputs. Furthermore, we fit the constant $c$ of the mean function, assigning it a $\mathcal{N}(0, 1)$ prior as well. In Tab.~\ref{tab:model}, we display the parameters of the MC approximations for various tasks. \textit{ No. $f$ } is the maximal number of functions used in the MC computation of the acquisition function. \textit{No. Reamples} is the number of initial posterior draws maximally used for the re-sampling of functions from the posterior $p(f|\rho)$. Lastly, . \textit{No. $\fopt$} is the number of optimal values used in the computation of \mname{}-\mes{}.
\begin{table}[htbp]

    \centering
    \begin{tabular}{c|c|c|c|c}
    \hline
    Task & No. $f$ & No. RFFs & No. Resamples  & No. $\fopt$  \\
    \hline
    Synthetic Good & $768$ & $2048$ & $1.5 * 10^5$ & $32$\\
    Synthetic Bad & $768$ & $2048$ & $1.5 * 10^5$ & $32$\\
    PD1 & $512$ & $4096$ & $2*10^5$ & $32$\\
    Appendix & $512$ & $1024$ & $10^5$ & $32$\\
    \hline
    \end{tabular}
    \caption{Budget-related parameters of the Monte Carlo approximations for all tasks.}
    \label{tab:model}
\end{table}
\subsection{Benchmarks}\label{app:benchmarks}
We outline the benchmarks used, their search spaces and the amount of synthetic noise added. When adding noise, we intend for the ratio of  noise variance to total output range to be approximately equal across benchmarks. 
\begin{table}[htbp]
    \centering
    \begin{tabular}{c|c|c|c}
    \hline
    Task & Dimensionality & $\sigma_\epsilon$& Search space  \\
    \hline
    Hartmann (4D) & $4$ & $0.25$ & $[0, 1]^D$\\
    Levy (5D) & $5$ & $0.5$ & $[-5, 5]^D$\\
    Hartmann (6D) & $6$ & $0.25$ & $[0, 1]^D$\\
    Rosenbrock (6D) & $6$ & $5$ & $[-2.048, 2.048]^D$\\
    Stybtang (7D) & $7$ & $1$ & $[-4, 4]^D$\\
    \hline
    \end{tabular}
    \caption{Benchmarks used for the Bayesian optimization experiments.}
    \label{tab:bench}
\end{table}
\subsection{Priors}\label{app:priors}

For synthetic benchmarks, the approximate optima of all included functions can be obtained in advance. Thus, the correctness of the prior is ultimately known in advance. For a function of dimensionality $d$ with optimum at $\xopt{}$, the well-located prior is constructed by sampling an offset direction $\bm{\epsilon}$ and scaling the offset by a dimensionality- and quality-specific term $c(d, q) = q\sqrt{d}$. For the well-located prior on synthetic tasks, we use $q=0.1$, which implies that the optimum is located 10\% of the distance across the search space away from the optimum,  and construct a Gaussian prior as

\begin{equation}
    \pi_{\xopt}(\bm{x}) \sim \mathcal{N}(\xopt + c_d\bm{\epsilon}/||\bm{\epsilon}||, \sigma_s), \quad \bm{\epsilon} \sim \mathcal{N}(0, \bm{I}).
\end{equation}
with $\sigma_s = 25\%$ for all tasks and prior qualities. For our 20 runs of the well-located prior, this procedure yields us 20 unique priors per quality type, with identical offsets from the true optimum. No priors with a mode outside the search space were allowed, such priors were simply replaced. For the misinformed priors, we set $q = 1$, guaranteeing that the mode of the prior will be outside of the search space, and subsequently relocating to the edge of the search space by its shortest path. Priors for all tasks are displayed in Tab.~\ref{tab:priors}. For the PD1 tasks, the location for the priors were obtained from MF-Prior-Bench(~\url{https://github.com/automl/mf-prior-bench}). However, these priors require offsetting in order to not be too strong, thus making subsequent BO obsolete. PD1 priors are provided in $[0, 1]$-normalized space for simplicity.

\begin{table}[htbp]
    \centering
    \begin{tabular}{c|c|c|c|c|c}
    \hline
    Task & Location & Offset, good & Offset, bad & $\sigma_s$ \\
    \hline
    Hartmann (4D) & $[0.19, 0.19, 0.56, 0.26]$ & $0.1\sqrt{D}$ & $\max$ & $0.25$\\
    Levy (5D) & $[1]^D$ & $1\sqrt{D}$ & $\max$ & $2.5$\\
    Hartmann (6D) & $ [0.20, 0.15, 0.48, 0.28, 0.31, 0.66]$ & $0.1\sqrt{D}$ & $\max$ & $0.25$\\
    Rosenbrock (6D) & $[1]^D$ & $0.4096\sqrt{D}$ & $\max$ & $1.024$\\
    Stybtang (7D) & $[-2.9]^D$ & $0.8\sqrt{D}$ & $\max$ & $2$\\
    PD1-WMT & $[0.90, 0.69, 0.02, 0.97]$ & $0.05\sqrt{D}$ & N/A & $0.25$\\
    PD1-CIFAR & $[1, 0.80, 0.0, 0.0]$ & $0.05\sqrt{D}$ & N/A & $0.25$\\
    PD1-LM1B & $[0.91, 0.67, 0.36, 0.85]$ & $0.05\sqrt{D}$ & N/A & $0.25$\\
    
    \hline
    \end{tabular}
    \caption{$\rho_{\bm{x}}^*$ for synthetic BO tasks of both prior qualities and PD1.}
    \label{tab:priors}
\end{table}

\section{Additional Experiments}
We provide complementary experiments to those introduced in the main paper. Firstly, we display results when \mname{} is used with a prior $\pi_{\fopt}$ over the optimal value in Sec.~\ref{app:maxvalue}. In Sec.~\ref{app:batch}, we demonstrate \mname{}:s extensibility to batch evaluations, seamlessly extending the work of~\citep{wilson2017reparam}.

\subsection{Synthetic Matern Kernel Experiments}~\label{sec:matern}
We evaluate \mname and all baselines on the synthetic tasks with a Matern-5/2 kernel and the good user belief over the optimum. We note that roughly half of all \pibo{} runs struggle with numerical instability from iteration 60 onwards, which produces stagnation in performance and infrequent gains. 
\begin{figure}[tbp]
  \centering
\includegraphics[width=\linewidth]{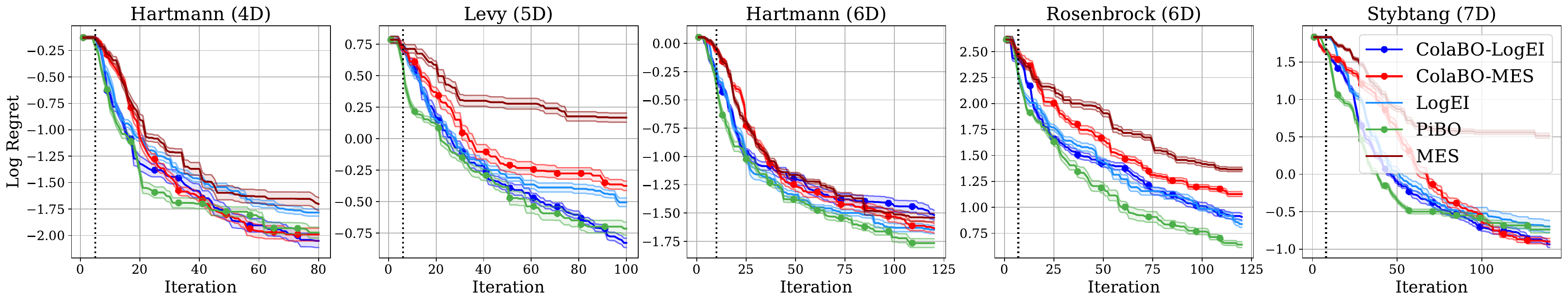} 
\caption{\mname{} on the synthetic tasks with a Matern kernel. Due to the difficulty of the RFF approximation, \mname{}-\logei{} struggles on Hartmann (6D), and \mname{} performance is marginally worse on aggregate.}
\label{fig:maxval}
\end{figure}
\subsection{Max-value Priors}\label{app:maxvalue}
We evaluate \mname{} with priors over the optimal value $\pi_{\fopt}$ in Figure~\ref{fig:maxval}. For each task, we place a Gaussian prior over the optimal value, centering it exactly at the optimal value. Notably, such a prior substantially influences the exploration-exploitiation trade-off; if the prior suggests that the incumbent has a value close to the optimal one, we are encouraged to exploit as samples with well-above-optimal values in exploratory will be discarded. Conversely, we are heavily encouraged to explore if the current best observation holds a value that we believe is far from optimal. On Hartmann (6D), we can see this behavior at play. Initial performance is poorer for \mname{} than their respective baselines, presumably due to above-average exploration, but terminal performance is better.
\begin{figure}[tbp]
  \centering
\includegraphics[width=\linewidth]{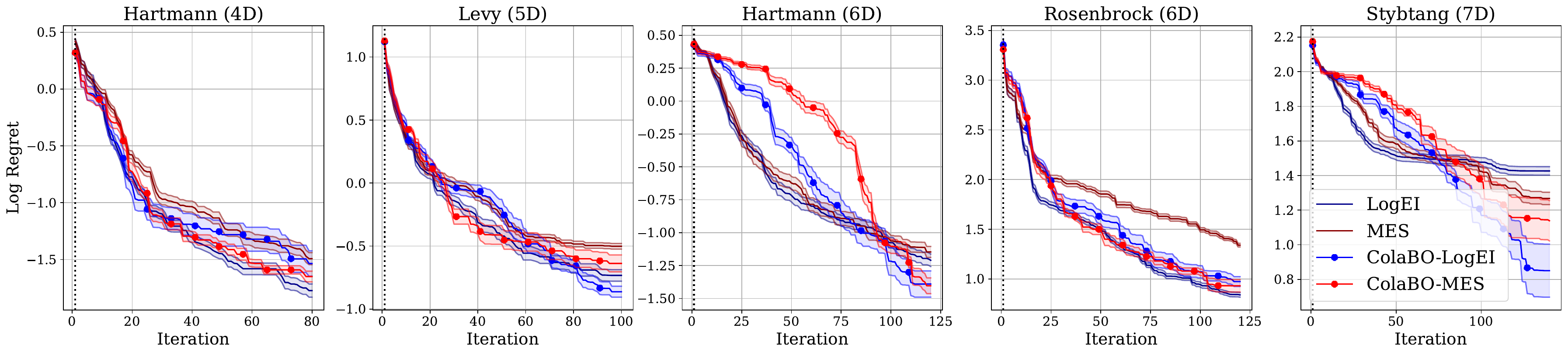} 
\caption{\mname{} with priors over the optimal value. Terminal performance substantially increases on 3 out of 5 benchmarks (Levy, Hartmann (6D), Stybtang), and is approximately preserved on the final two. \mname{}-\mes{} improves marginally more than \mname{}-\logei{} when utilizing a prior $\rho_{f_{\bm{x}}}^*$ over the optimal value.}
\label{fig:maxval}
\end{figure}
\subsection{Batch Evaluations}\label{app:batch}
We evaluate \mname{} on batch evaluations, utilizing the sequential greedy technique for MC acquisition functions from~\citet{NEURIPS2018_498f2c21}. Drop-off from sequential to batch evaluations is not evident from the plots, as ordering between sequential and batch varies with the benchmark. While unpredictable, we speculate that the altered exploration-exploitation trade-off provided by the batched acquisition function is occasionally beneficial in the presence of auxilliary user beliefs $\rho_{\bm{x}}^*$.
\begin{figure}[tbp]
  \centering
\includegraphics[width=\linewidth]{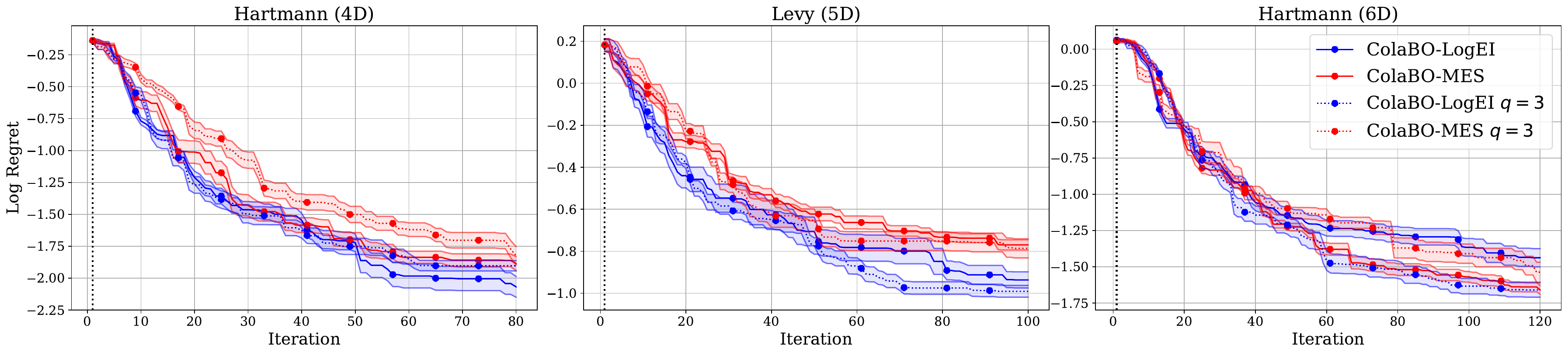} 
\caption{$q =1$ (sequential) and $q =3$ batch evaluation on a subset of synthetic functions with well-located priors for \mname{}-\logei{} and  \mname{}-\mes{}. Total function evaluations are plotted for both sequential and batched variants, leaving them with the same number of total function evaluations.} 
\label{fig:batch}
\end{figure}
\end{document}